\documentclass[final]{cvpr}

\usepackage{times}
\usepackage{epsfig}
\usepackage{graphicx}
\usepackage{subcaption}
\usepackage{amsmath}
\usepackage{amssymb}
\usepackage{mathrsfs}
\usepackage{colonequals}
\usepackage{enumitem}
\usepackage{mathtools}
\usepackage{tabularx}
\usepackage{appendix}
\usepackage{stfloats}
\usepackage[nohyperlinks, nolist]{acronym}

\DeclareMathOperator{\Tr}{Tr}

\usepackage[pagebackref=true,breaklinks=true,colorlinks,bookmarks=false]{hyperref}

\def\cvprPaperID{4675} 
\def\confYear{CVPR 2021}

\begin{acronym}[Acronyms]
\acro{nst}[NST]{Neural Style Transfer}
\acro{cmd}[CMD]{Central Moment Discrepancy}
\acro{da}[DA]{Domain Adaptation}
\acro{cnn}[CNN]{Convolutional Neural Network}
\acro{ot}[OT]{Optimal Transport}
\acro{mmd}[MMD]{Maximum Mean Discrepancy}
\acro{mgf}[MGF]{moment generating function}
\acro{rkhs}[RKHS]{reproducing kernel Hilbert space}
\acro{mm}[MM]{Moment Matching}
\end{acronym}

\title{In the light of feature distributions: moment matching for Neural Style Transfer}

\author{
Nikolai Kalischek$\qquad\qquad$
Jan D.\ Wegner$\qquad\qquad$
Konrad Schindler\\
EcoVision Lab, Photogrammetry and Remote Sensing, ETH Zürich\\
{\tt\small \{nikolai.kalischek,jan.wegner\}@geod.baug.ethz.ch, schindler@ethz.ch}
}

\begin{document}
\makeatletter
\newlength{\teaserht}
\setlength{\teaserht}{4.2cm}
\makeatother

\twocolumn[{%
\renewcommand\twocolumn[1][]{#1}%
\maketitle
\begin{center}
\vspace{-1em}
\setlength{\tabcolsep}{2pt}
\renewcommand{\arraystretch}{0.8}
\begin{tabular}{ccccc}
    \includegraphics[height=\teaserht]{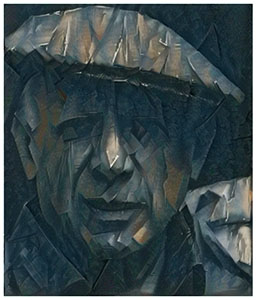} &
    \includegraphics[height=\teaserht]{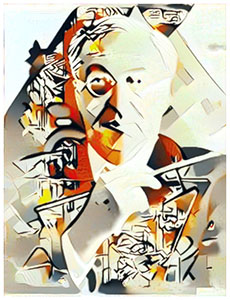} &
    \includegraphics[height=\teaserht]{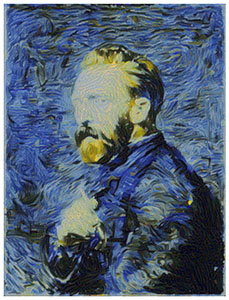} &
    \includegraphics[height=\teaserht]{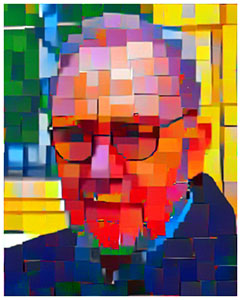} &
    \includegraphics[height=\teaserht]{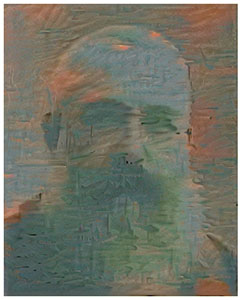} \\
    P.\ Picasso &
    W.\ Kandinsky &
    V.\ Van Gogh &
    G.\ Richter &
    C.\ Monet
\end{tabular}
\vspace{-0.5em}
\captionof{figure}{Style transfer results: artists' portraits rendered in their own painting style by our method.}\label{fig:artistinception}
\vspace{0.6em}
\end{center}%
\setlength{\tabcolsep}{6pt}
\renewcommand{\arraystretch}{1.0}
}]

\maketitle
\begin{abstract}
\vspace{-0.5em}
Style transfer aims to render the content of a given image in the graphical/artistic style of another image. The fundamental concept underlying \emph{Neural} Style Transfer (NST) is to interpret style as a distribution in the feature space of a Convolutional Neural Network, such that a desired style can be achieved by matching its feature distribution. We show that most current implementations of that concept have important theoretical and practical limitations, as they only partially align the feature distributions. We propose a novel approach that matches the distributions more precisely, thus reproducing the desired style more faithfully, while still being computationally efficient. Specifically, we adapt the dual form of \ac{cmd}, as recently proposed for domain adaptation, to minimize the difference between the target style and the feature distribution of the output image. The dual interpretation of this metric explicitly matches \emph{all} higher-order centralized moments and is therefore a natural extension of existing NST methods that only take into account the first and second moments. Our experiments confirm that the strong theoretical properties also translate to visually better style transfer, and better disentangle style from semantic image content.
\end{abstract}

\vspace{-0.5em}
\section{Introduction}
\vspace{-0.1em}
In 2017 \textit{Loving Vincent} was released, the first fully painted feature film with $>$65,000 frames. Indeed, every single frame is an oil painting drawn by one of over 100 artists. The creation of the movie was split into two steps. First, the entire movie was produced with real actors in front of a green screen, which was then replaced by Van Gogh paintings. In a second step, each frame was painted over by an artist with the techniques and style of Van Gogh, which took over six years to complete.

Attempts to automate this form of texture synthesis, termed \textit{style transfer}, date back to at least the mid-90s \cite{heeger1995pyramid}. More recently, Gatys \etal \cite{gatys2016image} pioneered the idea of \textit{\ac{nst}}. It is based on the idea that the deep layers of a pre-trained \ac{cnn} encode high-level semantic information and are insensitive to the actual appearance, whereas shallow layers learn low-level features such as color, texture and brush patterns.
A fundamental question that arises in this context is how to define style. Li \etal \cite{li2017demystifying} proved that the loss introduced in \cite{gatys2016image} can be rewritten as a \ac{mmd}, offering an interpretation of style transfer as aligning feature distributions. In fact, most existing methods can be interpreted in this way.
This has led to a series of works all centered around aligning feature distributions of \acp{cnn}, linking style transfer to \ac{da}.
Here we look deeper into that interpretation. By translating \ac{nst} to distribution matching, it becomes amenable to a suite of tools developed to measure the divergence between probability distributions, such as integral probability metrics, $f$-divergences and \ac{ot}. 

Divergences $d(P,Q)$ between two distributions, respectively probability measures, are in general not metrics, but they should fulfil the weaker conditions of \emph{(i)} non-negativity: $d(P,Q)\!\geq\!0$; and \emph{(ii)} identity of indiscernibles: $d(P,Q)\!=\!0\text{ iff }P\!=\!Q$.
However, \textit{in the light of feature distributions, existing style transfer methods suffer from rather elementary theoretical limitations}. 
Broadly, there are two schools. Either the distributions are unrestricted, but the discrepancy between them is measured without adhering to the law of indiscernibles \cite{gatys2016image, li2017demystifying, huang2017arbitrary, risser2017stable}; or the distributions are approximated roughly with simple functions, so that they admit closed-form solutions \cite{pmlr-v108-mroueh20a, kolkin2019style, li2017universal, lu2019closed}.
\vspace{0.5em}

Here, we show how to overcome these limitations with the help of the recently proposed framework of \acfp{cmd} \cite{zellinger2019robust}.
That (pseudo-)metric is based on the representation of distributions as moment sequences on compact intervals. In the limit, \ac{cmd} is an integral probability metric on the set of compactly supported distributions, so it complies with the law of indiscernibles (as well as non-negativity) by definition.
Importantly, in its dual formulation the \ac{cmd} is computationally efficient, and approximations can be seamlessly justified with an upper bound on the central moments \cite{zellinger2017central}.
In summary, we make the following contributions:
\emph{(i)}
We systematically categorize existing \ac{nst} methods according to their way of aligning distributions;
\emph{(ii)}
we make explicit underlying approximations and highlight the corresponding limitations;
\emph{(iii)} We propose a novel \ac{nst} algorithm based on the \acl{cmd}.
To our knowledge, our method is the first one that aligns style distributions in a rigorous and computationally efficient manner, with theoretically grounded approximation bounds.
Empirically, the method achieves a more perspicuous separation between artistic style and semantic content, and enables visually more compelling style transfer according to a user study with \textgreater50 participants.

\vspace{1em}
\section{Related work}
\vspace{0.5em}

\paragraph{Style Transfer} has been an active research topic in computer vision for at least two decades. Until recently it was based on hand-crafted features and styles. This includes stroke-based rendering \cite{kyprianidis2012state} to repaint an image with a set of brush strokes \cite{hertzmann1998painterly}, image quilting \cite{efros2001image} where texture is synthesized in small patches according to a segmentation map, or image analogies \cite{hertzmann2001image} that learn style filters in a supervised fashion. The shift to \acp{cnn} has given rise to \acl{nst}. Current \ac{nst} techniques can be categorized as being based on either image optimization or model optimization \cite{jing2019neural}. Methods in the first group iteratively transfer style to each new output image, following the seminal paper of \cite{gatys2016image}. That work first introduced the idea to match feature statistics of intermediate layers in a \ac{cnn}. Subsequent works explored different directions to improve the quality of stylization. Risser \etal \cite{risser2017stable} circumvent instabilities of the optimization by incorporating additional histogram and total variation losses.
To further enhance the preservation of low-level content such as edges, Li \etal \cite{li2017laplacian} add a Laplacian loss. In order to transfer style between semantically matching patches (\eg, from eyes of a dog to eyes of a cat), \cite{mechrez2018contextual} defines a loss that compares regions with similar semantic meaning. Similarly, \cite{li2016combining} use MRFs to find the nearest-neighbor patch in the feature space of the style image. Both require similar shapes and boundaries in the content and style images. Gatys \etal \cite{gatys2017controlling} also went on to add user-control for perceptual factors such as color or scale, \eg, by transferring style only in the luminance channel to preserve color. Recently, Kolkin \etal \cite{kolkin2019style} also incorporate user-defined spatial constraints, via appropriate weights in the cost function.
\vspace{0.5em}

Iterative optimization per image is comparatively slow. Model optimization methods instead employ feed-forward networks \cite{johnson2016perceptual,ulyanov2016instance} trained offline on large datasets, to achieve real-time style transfer.
Initially they were restricted to a fixed set of styles \cite{ulyanov2016texture, ulyanov2016instance, dumoulin2016learned, li2017diversified}. Later they were extended to handle unseen styles. Huang and Belongie \cite{huang2017arbitrary} propose an adaptive instance normalization layer that normalizes the content image with affine parameters from the style image, Chen and Schmidt \cite{chen2016fast} define a swap layer that replaces content feature patches with matching style feature patches. However, there is a price to pay for fast feed-forward inference, as it does not reach the quality of iterative methods. Recently it has been shown that
adaptive instance normalization, as well as the whitening color transform \cite{li2017universal} are special cases of an \ac{ot} map between Gaussian distributions, thus providing some theoretical foundation for feed-forward models \cite{lu2019closed,pmlr-v108-mroueh20a}.

\paragraph{Domain Adaptation}
is a particular instance of transfer learning, \ie, distilling and transferring knowledge across different domains. \acf{da} utilizes supervision in a source domain to guide the learning for a target domain where no labeled data is available \cite{csurka2017domain}. The principle is that the shift between the source and target domains can be measured, and therefore also minimized. Several authors have noted the close relation to \ac{nst} \cite{li2017demystifying, bousmalis2017unsupervised}.
A common approach is to learn a joint feature space by aligning the distributions in the latent feature space with measures such as Kullback-Leibler divergence \cite{zhuang2015supervised}, \acl{mmd} \cite{long2017deep} or correlation alignment \cite{sun2016deep}. Also related to style transfer, another approach to \ac{da} is to directly learn the mapping between the source and target  domains, \eg, using GANs \cite{bousmalis2017unsupervised}. For an overview of \ac{da}, see \cite{csurka2017domain,wang2018deep}. Here we make use of yet another idea originally aimed at \ac{da}, emphasizing its close relation to style transfer.

\section{Method}
We first briefly review the core ideas of \textit{\acl{nst}}. In that context, we revisit several existing methods and classify them into three categories. By taking the view of distribution alignment to its logical end, we then go on to provide an alternative loss function that has strong theoretical guarantees, is efficient to compute, and delivers visually appealing results (\cf Fig.~\ref{fig:artistinception}).

\subsection{Neural style transfer}

The fundamental idea of \ac{nst} is to use a pretrained, deep neural network to generate an image $I_o$ with the content-specific features of a content image $I_c$ and the style-specific features from a style image $I_s$. Typically, one minimizes a convex combination of a content and a style loss:
\begin{equation}\label{eq:nstloss}
    \mathcal{L} = \alpha \mathcal{L}_\text{content} + (1 - \alpha) \mathcal{L}_\text{style}.
\end{equation}
We further specify those losses following the notation of \cite{pmlr-v108-mroueh20a}.
Let $g$ be a deep encoder, say VGG-19 \cite{simonyan2014very}. For a specific layer $l$ with corresponding output feature map of spatial dimension ${H_l\cdot W_l=n_l}$ and channel depth $C_l$, we denote the $j$\textsuperscript{th} component of the feature map as a (reshaped) function ${F^l_j: \mathbb{R}^{d} \rightarrow \mathbb{R}^{C_l}}$, ${j\in [n_l]}$.
We write ${\mathbf{F}^l=(F^l_j)_{j\in [n]} \in \mathbb{R}^{C_l\times n_l}}$ and call $\mathbf{F}^l(I)$ the $l$\textsuperscript{th} (reshaped) \textit{feature map} of image $I$.
\Ie, the $L$\textsuperscript{th} feature map of image $I$ is the activation map after applying all layers $l=1,\dots, L$ to $I$.
Then, the content loss is proportional to 
\begin{equation}
    \mathcal{L}_\text{content}(I_o, I_c) \propto
    \sum_{l}
    || \mathbf{F}^l(I_o) - \mathbf{F}^l(I_c)||^2,
\end{equation}
where $l$ iterates over a set of layers of $g$. Commonly, only a single, deep layer is used to compute the content loss; whereas the style loss is an average over multiple layers, shallow and deep, with hyper-parameters $w_l$:
\begin{equation}
    \mathcal{L}_\text{style}(I_o, I_s) =
    \sum_l w_l    \mathcal{L}^l_\text{style}(I_o, I_s).
\end{equation}

\subsection{Style as feature distribution}\label{sec:stylefeaturedistribution}
Losses proposed for $\mathcal{L}^l_\text{style}$ can be categorized according to how they align distributions. We first need some additional definitions, again following \cite{pmlr-v108-mroueh20a}.
To obtain a distribution, we view the feature map $\mathbf{F}^l(I)$ as a $C_l$-dimensional empirical distribution measure over ${n_l=H_l\cdot W_l}$ samples. Note, by regarding the $n_l$ samples as an unordered set we explicitly discard the spatial layout.
This corresponds to the intuition that style attributes like color, strokes and texture are independent of the location. More formally, we define 
\begin{equation}
\nu^l: \mathbb{R}^d \rightarrow\mathscr{P}(\mathbb{R}^{C_l})
\quad,\quad
I \mapsto \frac{1}{n_l} \sum_{i=1}^{n_l}\delta_{F_j^l(I)},
\end{equation}
where $\mathscr{P}(\mathbb{R}^{C_l})$ is the space of empirical measures on $\mathbb{R}^{C_l}$. 
We abbreviate ${\nu_I^l = \nu^l(I)}$ and drop the layer index when not needed. With these definitions we now review existing style transfer methods in the light of distribution alignment.

\vspace{-1.0em}
\paragraph{MMD-based optimization.}
Already the first \ac{nst} paper \cite{gatys2016image} used statistics of feature maps to extract style-specific attributes of $I_s$, via the Gram matrix $G$. The Gram matrix contains 2\textsuperscript{nd}-order statistics, in our case correlations between corresponding channels in the feature map.
The link to aligning distributions may not be obvious, but Li \etal \cite{li2017demystifying} show that the style loss in \cite{gatys2016image} can be rewritten as an unbiased empirical estimate of the \ac{mmd} \cite{gretton2012kernel} with a polynomial kernel ${k(x,y) = (x^Ty)^2}$:
\begin{equation}\label{eq:mmd}
    \mathcal{L}^l_\text{style}(I_o, I_s) \propto \mathsf{mmd}^2[\mathbf{F}^l(I_o), \mathbf{F}^l(I_s)].
\end{equation}
Under the assumption that the \ac{rkhs} is \textit{characteristic} \cite{fukumizu2008kernel}, the \ac{mmd} vanishes if and only if the two distributions are the same. By treating the feature maps of $I_o$ and $I_s$ as samples, minimizing the objective (\ref{eq:mmd}) is the same as minimizing the discrepancy between $\nu_{I_o}$ and $\nu_{I_s}$.

\vspace{-1.0em}
\paragraph{Moment-based optimization}$\!\!$approaches explicitly minimize the difference between style distributions. Theoretical support for these methods comes from \acp{mgf}. It is known that a distribution is uniquely characterized by its moments if the \ac{mgf} is finite in an open interval containing zero. Hence, if two distributions with finite \ac{mgf} have equal moments, they are identical.

Besides relating style transfer to distribution alignment, Li \etal \cite{li2017demystifying} also introduced a style loss based on \emph{batch normalization} statistics. That loss is the first to explicitly match moments in feature space, namely the means $\mu_{\mathbf{F}^l(I)}$ and the standard deviations $\sigma_{\mathbf{F}^l(I)}$:
%
%
\begin{equation}
        \mathcal{L}^l_\text{style}(I_o, I_s) \propto\!\sum\limits_{i=1}^{C_l}[ (\mu^i_{\mathbf{F}^l(\!I_o\!)}\!\!-\! \mu^i_{\mathbf{F}^l(\!I_s\!)})^2 
        \!\!+\! (\sigma^i_{\mathbf{F}^l(\!I_o\!)}\!\!-\! \sigma^i_{\mathbf{F}^l(\!I_s\!)})^2].
\end{equation}
 Interestingly, moment alignment can also produce reasonable results when applied in feed-forward mode, without iterative optimization. Based on ideas from \cite{ulyanov2016instance,dumoulin2016learned}, Huang and Belongie \cite{huang2017arbitrary} align the mean and variance with a transformation layer. In summary, matching the mean and variance of the content image's feature space to that of the style image reduces the divergence between $\nu_{I_o}$ and $\nu_{I_s}$ -- but discrepancies due to higher-order moments remain.

\vspace{-1.0em}
\paragraph{Optimal Transport-based optimization}$\!\!$provides a principled framework to minimize the discrepancy between distributions, notably taking into account the geometry of the underlying spaces. When working in the space of probability measures ${\mathcal{P}_p(\mathbb{R}^d)}$ with bounded $p$\textsuperscript{th} moment, the \emph{Wasserstein distance} for ${P,Q \in \mathcal{P}_p(\mathbb{R}^d)}$ is defined as
\begin{equation}\label{eq:wasserstein}
    \mathcal{W}_p(P, Q)^p = \inf\limits_{\Gamma(P, Q)} \int ||x-y||^p d\pi(x,y).
\end{equation}
We can use the Wasserstein distance for back-propagation to minimize the discrepancy between $\nu_{I_o}$ and $\nu_{I_s}$. In general, computing the \ac{ot} has complexity ${O(n_l^3\log n_l)}$ and is not suitable for iterative optimization schemes.
However, restricting the distributions to Gaussians, ${\Tilde{\nu}_{I_o} \colonequals \mathcal{N}(\mu_{\nu_{I_o}},\Sigma_{\nu_{I_o}})}$ and ${\Tilde{\nu}_{I_s} \colonequals \mathcal{N}(\mu_{\nu_{I_s}},\Sigma_{\nu_{I_s}})}$
admits a closed form solution,
\begin{equation}
    \begin{split}
        \mathcal{W}_2(\Tilde{\nu}_{I_o},& \Tilde{\nu}_{I_s})^2= \|\mu_{\nu_{I_o}}\!\!- \mu_{\nu_{I_s}}\|^2_2\\ 
        &+ \Tr\big(\Sigma_{\nu_{I_o}}\!\!+ \Sigma_{\nu_{I_s}}\!\!- 2(\Sigma_{\nu_{I_o}}^\frac{1}{2}\Sigma_{\nu_{I_s}}\Sigma_{\nu_{I_o}}^\frac{1}{2})^\frac{1}{2}\big).
    \end{split}
    \label{eq:ot_approx}
\end{equation}
This is similar to matching the first and second moments as in moment-based optimization (higher-order moments of Gaussians are constant \wrt mean and variance). Conveniently, the \ac{ot} map can also be directly derived.
If one is willing to accept the Gaussian approximation, the style features can be aligned by iteratively minimizing $\mathcal{W}_2$, or by integrating the \ac{ot} map into the encoder-decoder network \cite{pmlr-v108-mroueh20a, kolkin2019style, lu2019closed, li2017universal}.
It has been shown \cite{pmlr-v108-mroueh20a,lu2019closed} that adaptive instance normalization can be seen as \ac{ot} of Gaussians with diagonal covariances.

\begin{figure*}[!tb]
    \centering
    \begin{subfigure}[b]{0.162\linewidth}
        \centering
        \includegraphics[width=\textwidth, trim=6 3 2 0, clip]{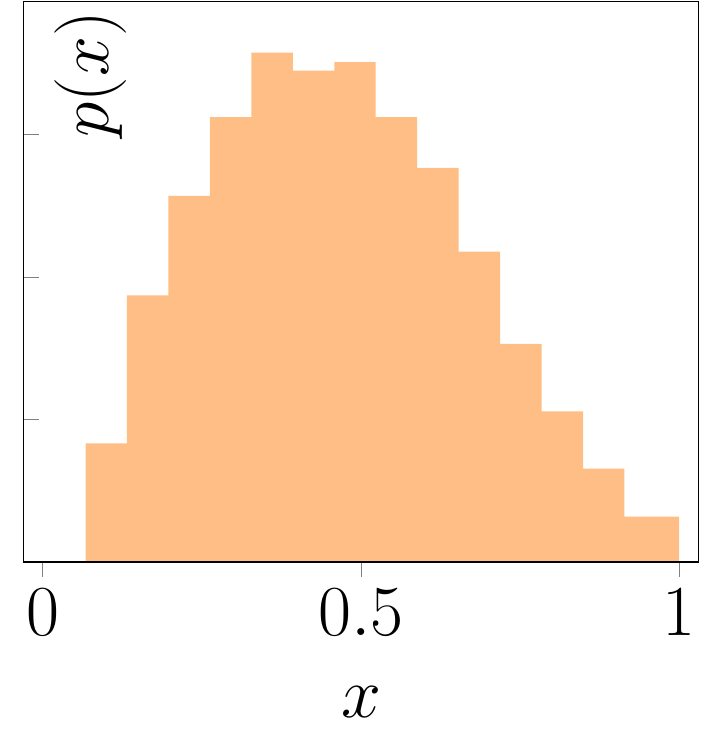}
        \caption{Source}
    \end{subfigure}
    \begin{subfigure}[b]{0.162\linewidth}
        \centering
        \includegraphics[width=\textwidth, trim=6 3 2 0, clip]{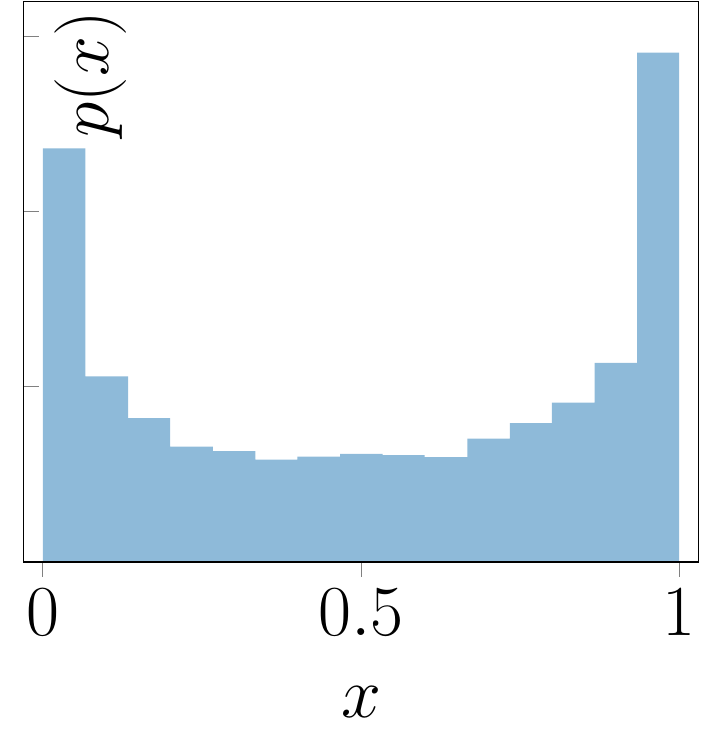}
        \caption{Target}
    \end{subfigure}
    \begin{subfigure}[b]{0.162\linewidth}
        \centering
        \includegraphics[width=\textwidth, trim=6 3 2 0, clip]{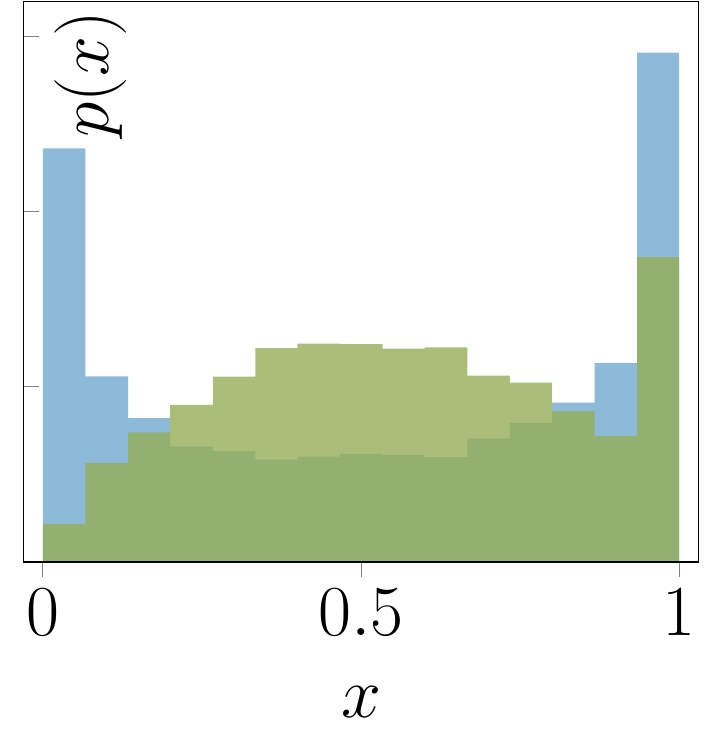}
        \caption{MMD}
    \end{subfigure}
    \begin{subfigure}[b]{0.162\linewidth}
        \centering
        \includegraphics[width=\textwidth, trim=6 3 2 0, clip]{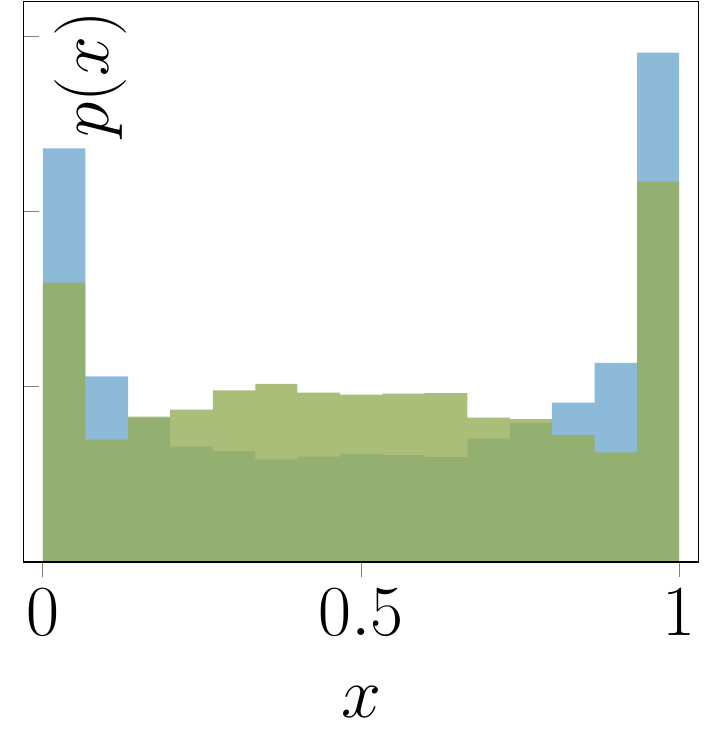}
        \caption{MM / OT}
    \end{subfigure}
    \begin{subfigure}[b]{0.162\linewidth}
        \centering
        \includegraphics[width=\textwidth, trim=6 3 2 0, clip]{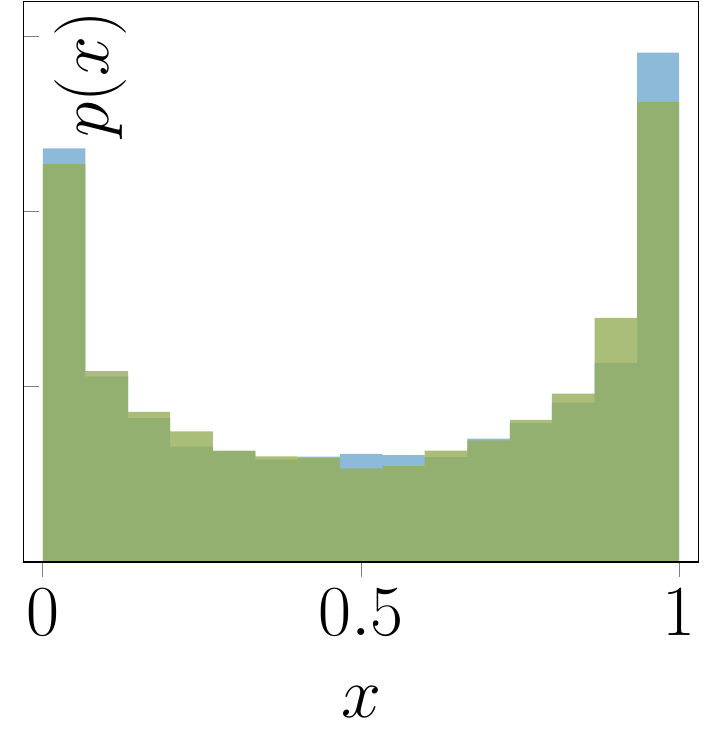}
        \caption{CMD, $K\!=\!5$}
    \end{subfigure}
    \begin{subfigure}[b]{0.162\linewidth}
        \centering
        \includegraphics[width=\textwidth, trim=6 3 2 0, clip]{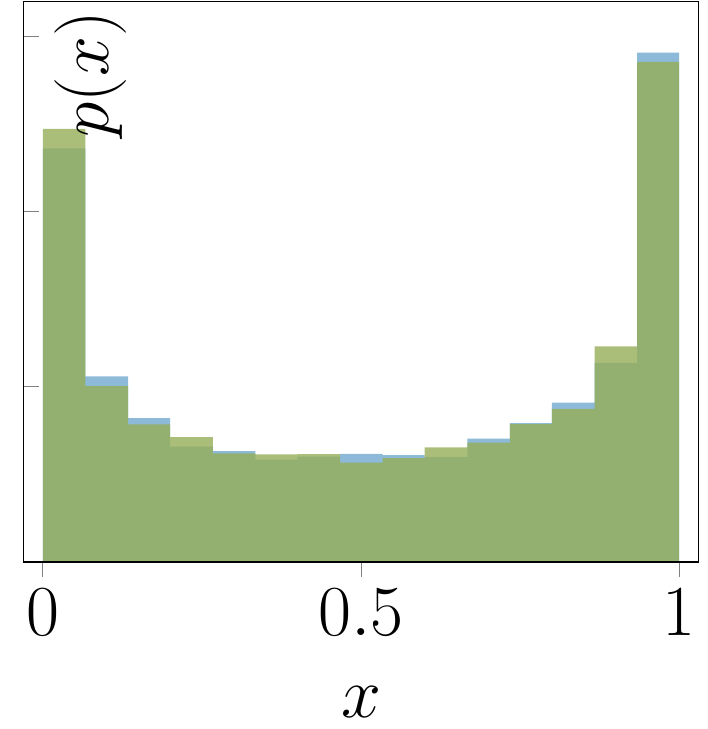}
        \caption{CMD, $K\!=\!50$}
    \end{subfigure}
    \caption{Illustration of distribution matching in 1D. The source $\sim\!Beta(2,3)$ and target $\sim\!Beta(0.5,0.45)$ cannot be aligned with MMD, MM or OT (which in 1D is the same as MM). On the contrary, CMD aligns them well already with five moments, and the residual error decreases asymptotically as more moments are added. See text for details.}
\label{fig:distmatching}
\end{figure*}

\vspace{0.7em}
\subsection{Motivation}\label{sec:motivation}
From a statistical perspective all three categories of methods contradict, to some extent, the goal of optimally aligning feature distributions.

Methods based on \textbf{\ac{mmd}} rely on simplistic (typically, linear or quadratic) kernels \cite{gatys2016image, li2017demystifying}. Previously, \cite{risser2017stable} already identified instabilities during training, as different distributions result in the same \ac{mmd}. They point out that changes in mean and variance can compensate each other, giving rise to the same Gram matrix (and thus the same \ac{mmd} with quadratic kernel), since the Gram matrix is related to \textit{non-central} second moments. 
We offer an alternative explanation why the Gram matrix violates the identity of indiscernibles: the quadratic kernel is non-characteristic, \ie, the map ${p\rightarrow \mathbb{E}_{x\sim p}[k(x, \cdot)]}$ is not injective and the distribution $p$ has no unique embedding in the \ac{rkhs}. %
Moreover, the quadratic kernel (resp.\ Gram matrix) is obviously restricted to 2\textsuperscript{nd} moments. It is highly unlikely that those are sufficient statistics for deep feature activations, so ${MMD(p,q)\!=\!0}$ almost certainly does not imply ${p\!=\!q}$.

A similar argument can be made about existing methods based directly on \textbf{\ac{mm}}, since they match only the means and variances. It is trivial to define two distinct distributions with the same variances -- \eg, a Gaussian ${\mathcal{N}(0, \sqrt{2})}$ and a Laplace distribution ${\mathcal{L}(0, 1)}$.

While \textbf{\ac{ot}} is a powerful framework at the conceptual level, it is hobbled by high computation cost.
The Gaussian approximation makes \ac{ot} tractable, but at the cost of losing information.
There is no evidence that the distributions $\nu_{I_o}$ and $\nu_{I_s}$ are (approximately) Gaussian -- in fact it is very unlikely, unless one artificially constrains them, thus seriously restraining the deep network's expressive power.
We claim that \ac{ot}, at least in its prevalent, restricted form, also mostly reduces to matching the first and second moments -- the approximations in~\eqref{eq:ot_approx} are completely defined in terms of means and covariances.

Finally, we point out the \textit{mean over-penalization} effect: \cite{zellinger2019robust} found instabilities of distribution alignment during \ac{da} training under small perturbations, which arise from the use of raw instead of centralized moments (as in \ac{mmd} with standard polynomial kernel and non-centralized integral probability metrics).
For details, please refer to \cite{zellinger2019robust}.


\vspace{0.7em}
\subsection{CMD for neural style transfer}
Instead of only matching first- and second-order moments, we propose to make use of a suitable integral probability metric, the \acl{cmd} \cite{zellinger2017central}. At its core, that metric utilizes the dual representation of compactly supported distributions as moment sequences. The translation to central moments leads to natural geometric relations such as variance, skewness and kurtosis.
Not that the idea of matching higher moments has been investigated in early work on texture synthesis~\cite{portilla2000parametric}, but so far has been disregarded in \ac{nst}.

In Fig.~\ref{fig:distmatching}, we illustrate the enhanced expressive power of \ac{cmd}.
In our toy example, the source and target are univariate $Beta$-distributions with different parameters, \ie, their third and fourth moments are non-zero. We represent each distribution with 10,000 samples and minimize the respective alignment loss with gradient descent. The example confirms that none of the three approaches based on first and second moments can align the two distributions (note that for the 1D case \ac{mm} and \ac{ot} are identical). On the contrary, \ac{cmd} aligns them nicely. 

The \ac{cmd} between two compactly supported distributions $P$ and $Q$ is defined as follows \cite{zellinger2019robust}:
\begin{equation}\label{eq:cmd}
\begin{split}
    \mathsf{cmd}_k(P,Q) \coloneqq& \sum\limits_{i=1}^k a_i \|c_i(P) - c_i(Q)\|_2\quad\text{, where}\\
    c_i(X)=&\begin{cases}
    \mathbb{E}_X[x] & i=1\\
    \mathbb{E}_X[\eta^{(i)}(x - \mathbb{E}_X[x])] &i\geq 2
    \end{cases}
    \end{split}
\end{equation}
with $a_i\geq 0$.  The $\eta^{(i)}(x)$ are monomial vectors of order $i$ defined as 
\begin{equation}
    \begin{split}
        \eta^{(i)}: &\mathbb{R}^m \rightarrow \mathbb{R}^{\frac{(i+1)^{m-1}}{(m-1)!}}\\
        &x \mapsto \big(x_1^{r_1}\dotsm x_m^{r_m}\big)_{\genfrac{}{}{0pt}{1}{(r_1,\cdots,r_m)\in \mathbb{N}_0^m}{r_1+\cdots+r_m=k}}.
    \end{split}
\end{equation}
By construction the \ac{cmd} is non-negative, respects the triangle inequality, and if ${P=Q}$ then ${\mathsf{cmd}_k(P,Q) = 0}$. Furthermore, \cite[Theorem~1]{zellinger2017central} states that ${\mathsf{cmd}_k(P,Q) = 0}$ implies ${P=Q}$ for ${k\rightarrow\infty}$, so \ac{cmd} is a metric on compactly supported distributions.

For practical applications computing $\mathsf{cmd}_\infty$ is obviously not possible, and we have to bound $k$ to ${K\!<\!\infty}$ from above. Compared to other approximations used for style transfer \cite{pmlr-v108-mroueh20a, kolkin2019style}, the bounded $\mathsf{cmd}_K$ has a natural theoretical justification. It can be shown \cite[Proposition~1]{zellinger2019robust} that the $i$\textsuperscript{th} term in the summation of equation \ref{eq:cmd} is bounded by an upper bound that strictly decreases with the order $i$. \Ie, the contribution of higher-order moment terms in equation (\ref{eq:cmd}) converges monotonically to $0$. To keep the implementation efficient we only compute the marginal moments, by restricting the monomial vectors to ${\eta^{(i)}(x) = (x_1^i,\cdots,x_m^i)}$.

\begin{figure*}[!tb]
 \setlength{\tabcolsep}{0pt}
\renewcommand{\arraystretch}{0.4}
   \begin{center}
        \begin{tabularx}{\textwidth}{*7{>{\centering\arraybackslash}X}}
            \includegraphics[width=0.14\textwidth]{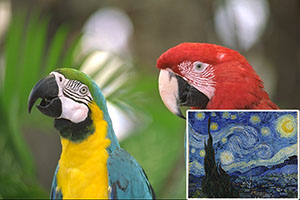}
            & \includegraphics[width=0.14\textwidth]{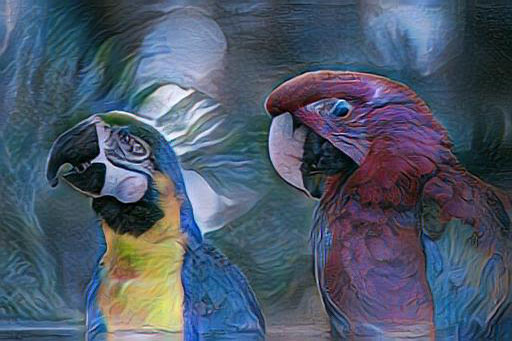}
            & \includegraphics[width=0.14\textwidth]{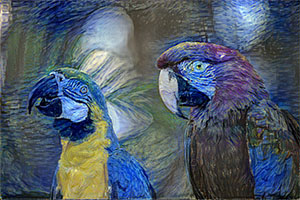}  
            & \includegraphics[width=0.14\textwidth]{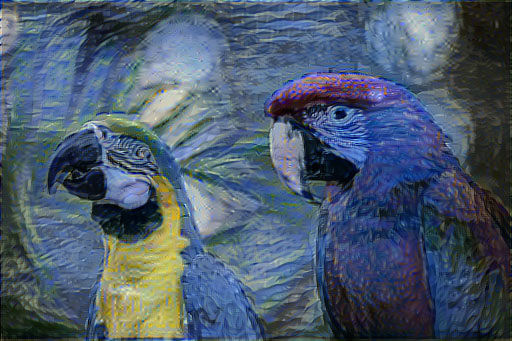} 
            & \includegraphics[width=0.14\textwidth]{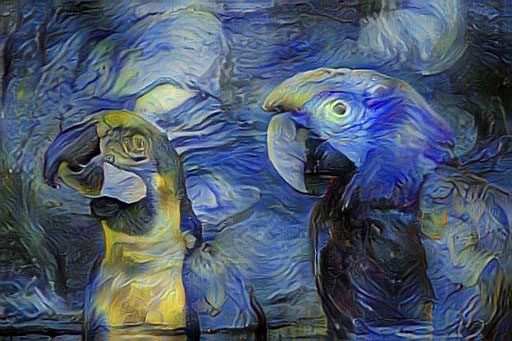}   
            & \includegraphics[width=0.14\textwidth]{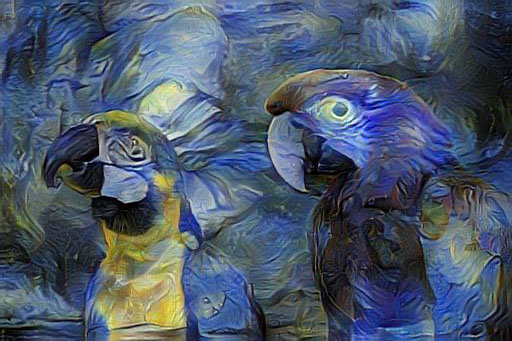} 
            & \includegraphics[width=0.14\textwidth]{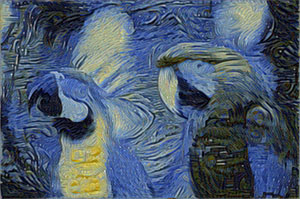}\\ 
            \includegraphics[width=0.14\textwidth]{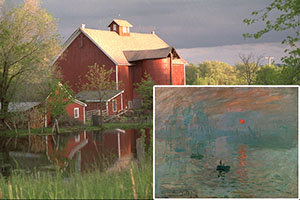}
            & \includegraphics[width=0.14\textwidth]{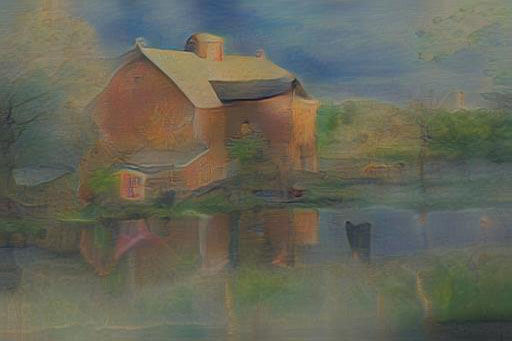}
            & \includegraphics[width=0.14\textwidth]{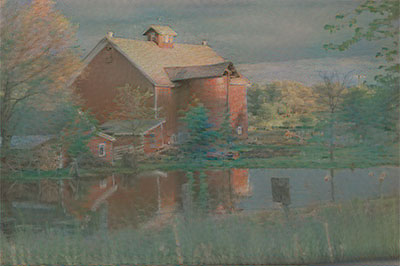}  
            & \includegraphics[width=0.14\textwidth]{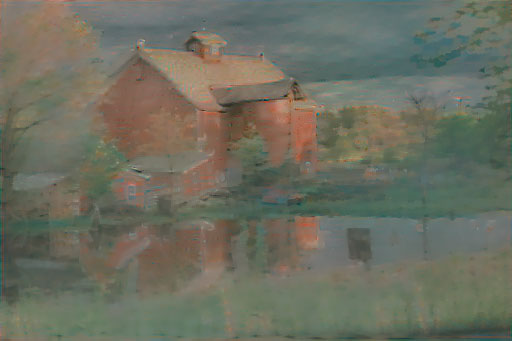} 
            & \includegraphics[width=0.14\textwidth]{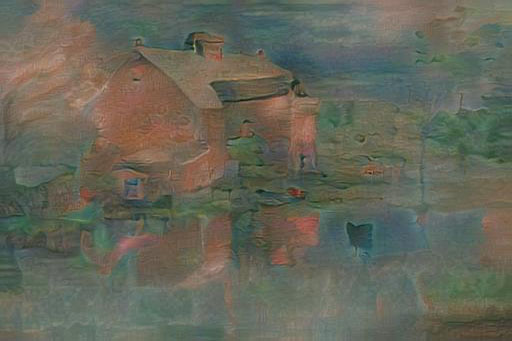}   
            & \includegraphics[width=0.14\textwidth]{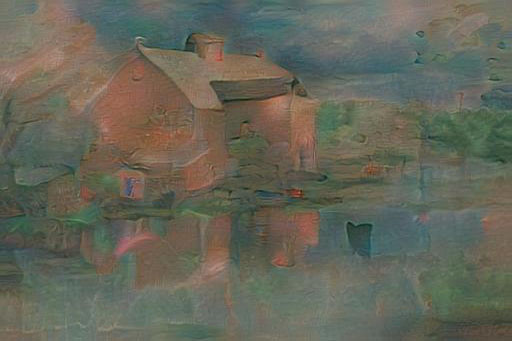} 
            & \includegraphics[width=0.14\textwidth]{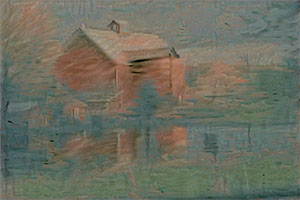}\\ 
            \includegraphics[width=0.14\textwidth]{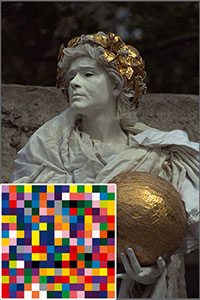}
            & \includegraphics[width=0.14\textwidth]{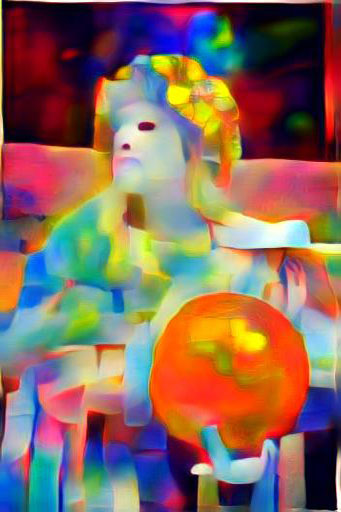}
            & \includegraphics[width=0.14\textwidth]{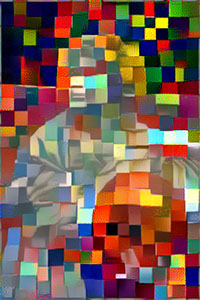}
            & \includegraphics[width=0.14\textwidth]{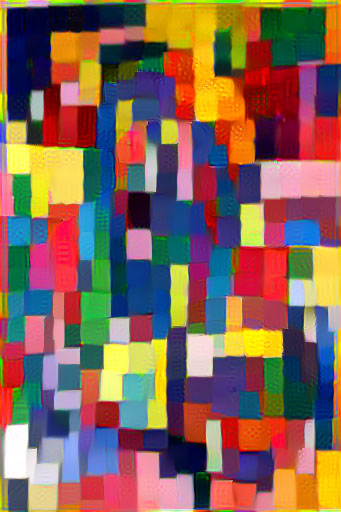} 
            & \includegraphics[width=0.14\textwidth]{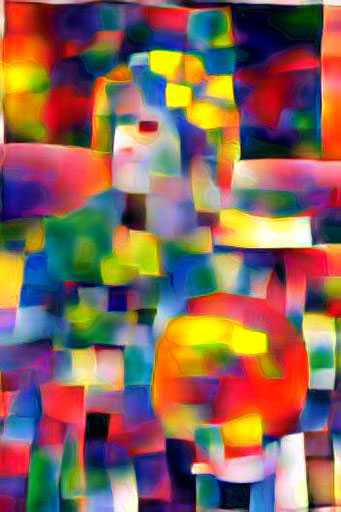}   
            & \includegraphics[width=0.14\textwidth]{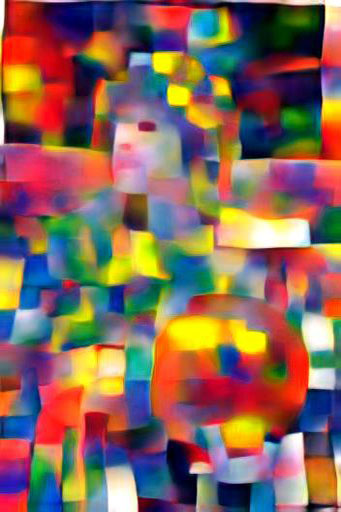} 
            & \includegraphics[width=0.14\textwidth]{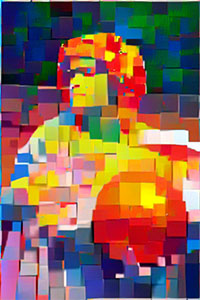}\\ 
            \includegraphics[width=0.14\textwidth]{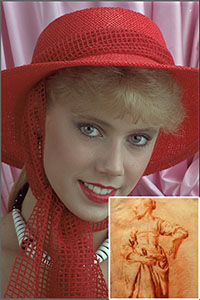}
            & \includegraphics[width=0.14\textwidth]{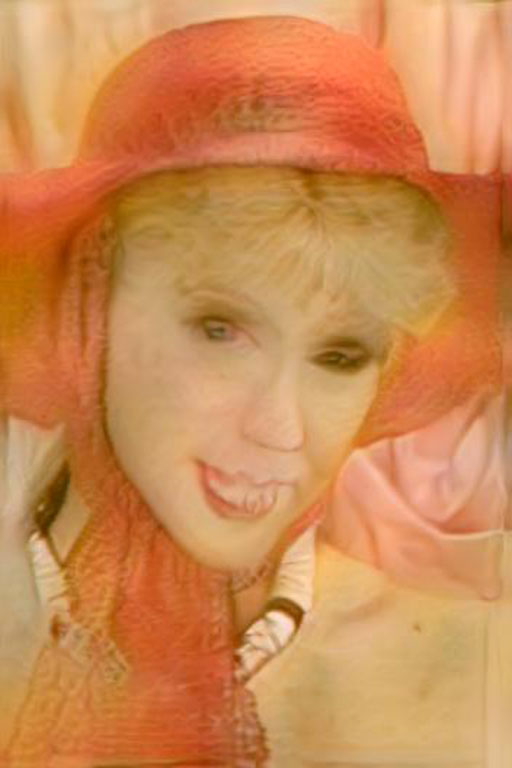}
            & \includegraphics[width=0.14\textwidth]{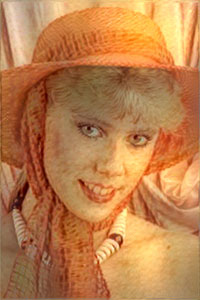}  
            & \includegraphics[width=0.14\textwidth]{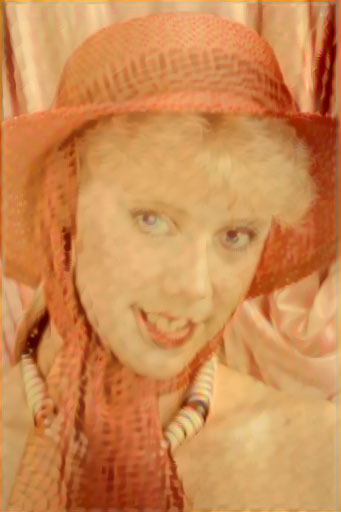} 
            & \includegraphics[width=0.14\textwidth]{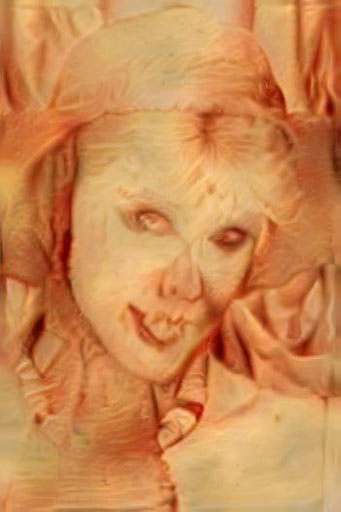}   
            & \includegraphics[width=0.14\textwidth]{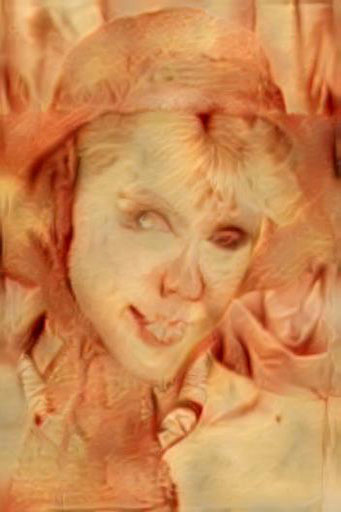}
            & \includegraphics[width=0.14\textwidth]{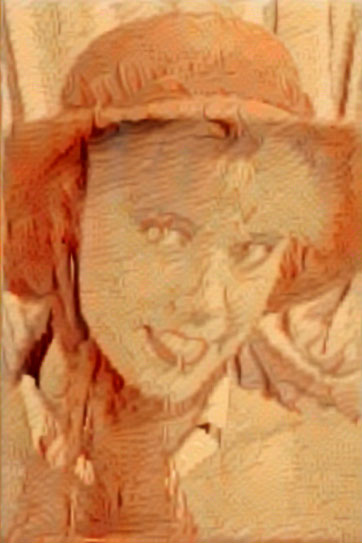}\\ 
            \includegraphics[width=0.14\textwidth]{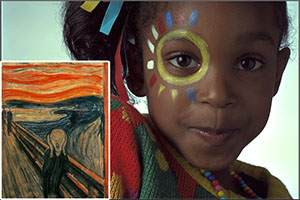}
            & \includegraphics[width=0.14\textwidth]{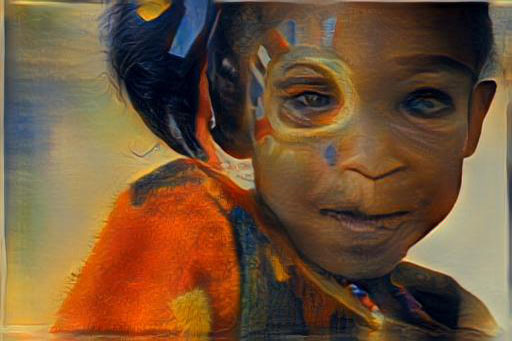}
            & \includegraphics[width=0.14\textwidth]{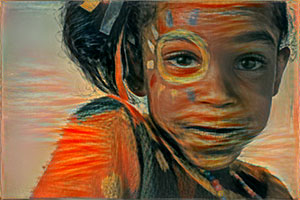}  
            & \includegraphics[width=0.14\textwidth]{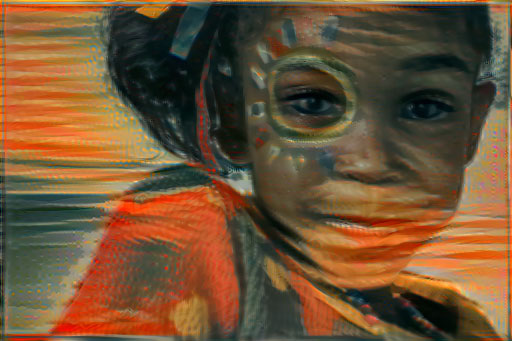} 
            & \includegraphics[width=0.14\textwidth]{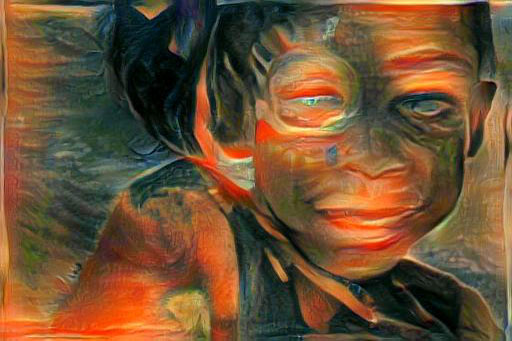}   
            & \includegraphics[width=0.14\textwidth]{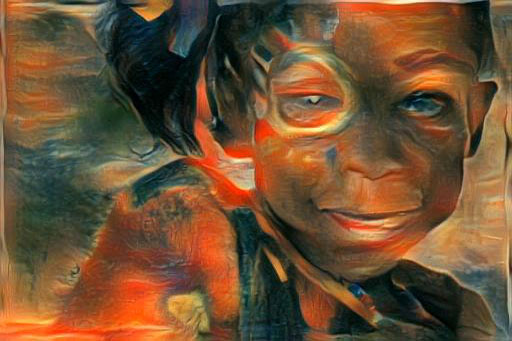} 
            & \includegraphics[width=0.14\textwidth]{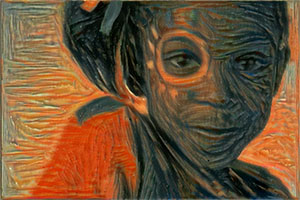}\\ 
            \includegraphics[width=0.14\textwidth]{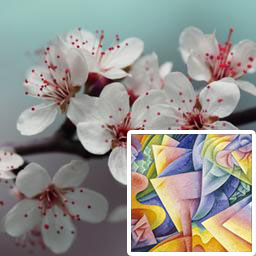}
            & \includegraphics[width=0.14\textwidth]{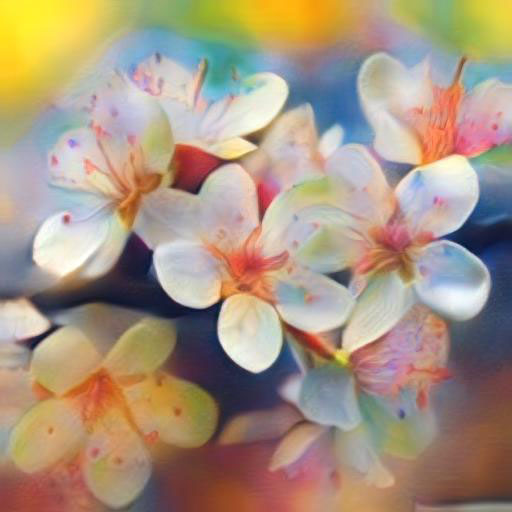}
            & \includegraphics[width=0.14\textwidth]{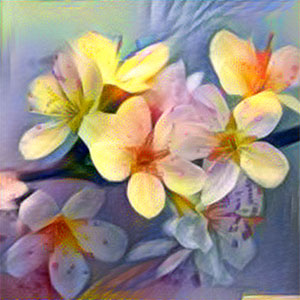}  
            & \includegraphics[width=0.14\textwidth]{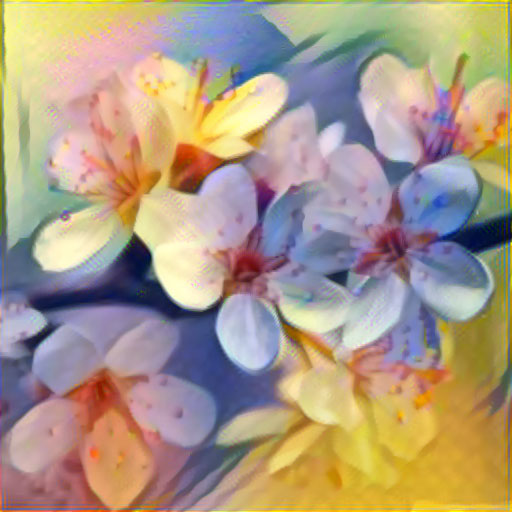} 
            & \includegraphics[width=0.14\textwidth]{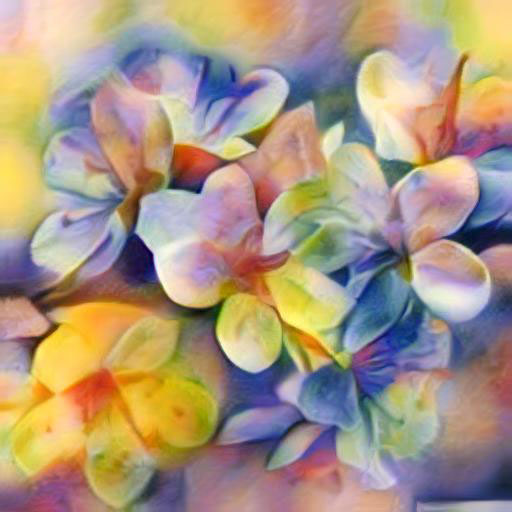}   
            & \includegraphics[width=0.14\textwidth]{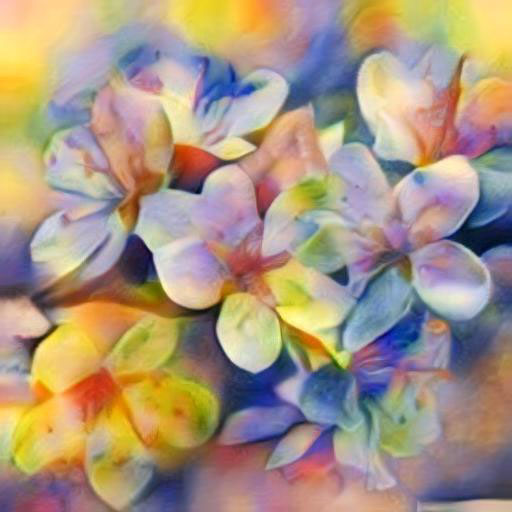} 
            & \includegraphics[width=0.14\textwidth]{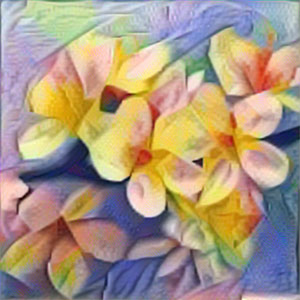}\\
            \includegraphics[width=0.14\textwidth]{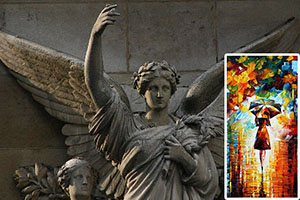}
            & \includegraphics[width=0.14\textwidth]{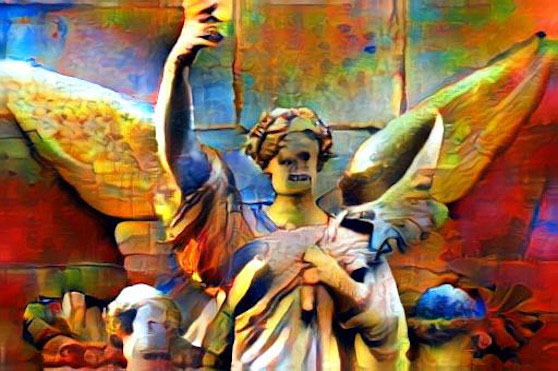}
            & \includegraphics[width=0.14\textwidth]{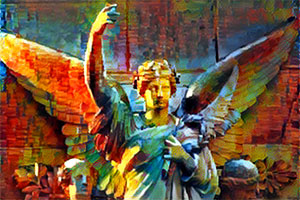}  
            & \includegraphics[width=0.14\textwidth]{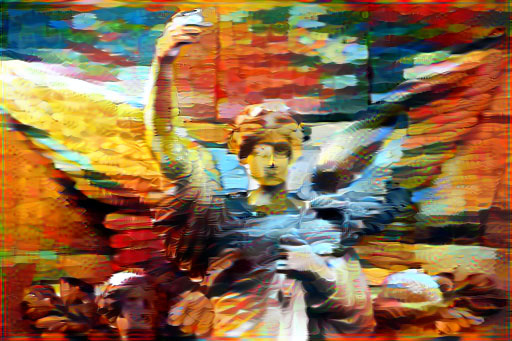} 
            & \includegraphics[width=0.14\textwidth]{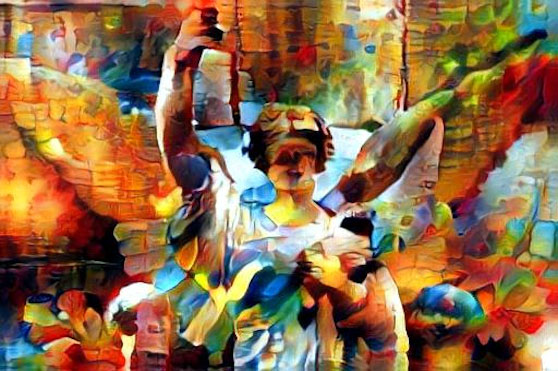}   
            & \includegraphics[width=0.14\textwidth]{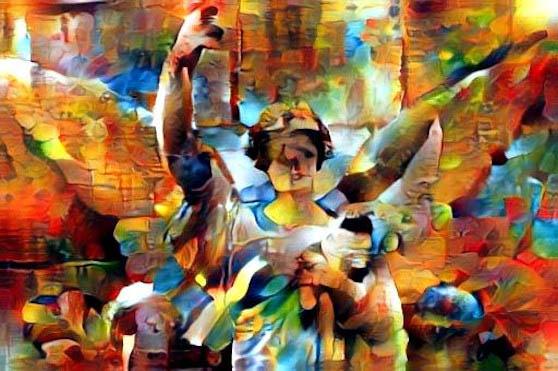} 
            & \includegraphics[width=0.14\textwidth]{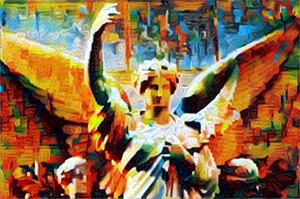}\\
            \rule{0pt}{14pt}\small (a) Input 
            & \rule{0pt}{12pt}\small (b) AdaIN \cite{huang2017arbitrary}
            & \rule{0pt}{12pt}\small (c) Gatys \cite{gatys2016image}
            & \rule{0pt}{12pt}\small (d) MM \cite{li2017demystifying}
            & \rule{0pt}{12pt}\small (e) OST \cite{lu2019closed}
            & \rule{0pt}{12pt}\small (f) WCT \cite{li2017universal}
            & \rule{0pt}{12pt}\small (g) Ours
        \end{tabularx}
    \end{center}
\vspace{-1.5em}
\caption{Style transfer results of our algorithm and of previous methods from all three categories. Best viewed on screen. Please zoom in to appreciate style details.}\label{fig:user-study-results}
\end{figure*}

Adapting \ac{cmd} to our style feature distributions is straight-forward. 
To fulfill the requirements, we wrap a sigmoid function $\sigma(\cdot)$ around each feature output so as to restrict the support of the empirical distribution to ${[0, 1]}$. 
With a slight abuse of notation we write $\sigma(\nu^l)$ for the $\nu^l$ computed from sigmoid-transformed features and define
\begin{equation}
    \mathcal{L}^l_\text{style}(I_o, I_s) \coloneqq \mathsf{cmd}_k\big(\sigma(\nu^l_{I_o}),\sigma(\nu^l_{I_s})\big),
    \label{eq:cmd_features}
\end{equation}
for layer $l$. The moments are simply the moments of the empirical measure, \ie powers of  ${\mathbb{E}[\mathbf{F}^l(I) - \mu_{\mathbf{F}^l(I)}] \in \mathbb{R}^{C_l}}$.
By adopting \ac{cmd} we have an integral probability metric for \ac{nst} at our disposal that not only has favourable theoretical properties, but is also easy to implement, computationally efficient, and able to handle complex feature distributions with significant higher-order moments.

\section{Results}

In this section, we compare our results with existing methods from each of the categories. After summarizing details of the implementation, we qualitatively evaluate the effects of aligning the style features with \ac{cmd}. Beyond visual comparisons, we report quantitative results from an user-study, which supports our hypothesis that higher-order moments carry important style information and should not be ignored. Lastly, we further investigate the impact of  different moments in an ablation study. 
\subsection{Experimental setup}

We employ VGG-19 \cite{simonyan2014very} as feature encoder and read out feature maps at layer levels ${l\in \{1\_1,2\_1,3\_1,4\_1,5\_1\}}$. Deviating slightly from the commonly used \ac{nst} setting, we work with the raw convolution outputs $\textit{conv-l}$ rather than their rectified versions $\textit{relu-l}$, since we clamp them to ${[0, 1]}$ with sigmoid activations for computing the \ac{cmd}, see~\eqref{eq:cmd_features}.
The content loss is computed on $\textit{conv4\_1}$, for the individual layers in the style loss we use the same weighting scheme as proposed in \cite{gatys2016image}.
Optimization is performed with Adam \cite{kingma2014adam}. Instead of blindly stopping after a fixed number of iterations, we implement a stopping criterion based on the difference of the current style loss and a moving average of the style loss.
We compare our algorithm to five baselines: one from the \ac{mmd} group \cite{gatys2016image}, two based on direct moment differences \cite{li2017demystifying,huang2017arbitrary} and two based on \ac{ot} \cite{li2017universal,lu2019closed}. We use the existing open-source implementations%
\footnote{For \cite{gatys2016image, li2017demystifying,lu2019closed}, original implementations by the authors; for \cite{huang2017arbitrary,li2017universal}, implementation provided by the authors of \cite{lu2019closed}.} %
and keep all hyper-parameters as proposed in the original papers, respectively source codes. Our implementation is based on PyTorch \cite{paszke2019pytorch} and is also publicly available.\footnote{Code: \href{https://github.com/D1noFuzi/cmd_styletransfer}{https://github.com/D1noFuzi/cmd\_styletransfer}} For our experiments we bound the order of the moments to ${K\!=\!5}$, as higher orders have little influence. 

\subsection{Qualitative results}

We have pinpointed theoretical limitations of previous \ac{nst} methods in Sec.~\ref{sec:motivation}. To see how these translate to concrete visual differences, we analyze how well the stylized images preserve three different style attributes, \textit{color, texture and stroke, shape}. See Fig.~\ref{fig:user-study-results}, and further results in the supplementary material.

\paragraph{Color and brightness.}
This paper is concerned with fully automatic \ac{nst}, without additional user control. Hence, the output should have the color palette of the style image. \Ie, only the semantic content of the content image should be retained, but colors should be replaced by those representative of the style, and in particular the two color spaces should not be mixed.
Looking at the 1\textsuperscript{st} row of Fig.~\ref{fig:user-study-results}, the red of the right parrot strongly leaks into the results of AdaIN, Gatys and \ac{mm}, and traces are also visible in WCT. Besides our method, those based on \ac{ot} fare best in terms of color palette, but \ac{ot} has a tendency towards exaggerated brightness variations not warranted by the content, \eg, the girl's face in row 5 and the background in row 6.
Indeed, it appears that local color and intensity information is to some degree hidden in higher-order moments. That observation is also supported by the ablation study in Sec.~\ref{sec:ablationstudies}. 
\vspace{-1.5em}
\paragraph{Texture and stroke.}
Maintaining strokes and textures is especially important when it comes to artistic style transfer, to preserve the concomitant individual painting techniques. We find that the proposed \ac{cmd} method is particularly good at replicating granular canvas, oriented brush strokes, \etc. Clear cases in point are rows 1 and 5 of Fig.~\ref{fig:user-study-results}, as well as the reflections on the lake in row 2. We also point out the particularly challenging example in the 4\textsuperscript{th} row. Zooming in on the style image, we can see the rough texture of the paper, as well as a preference for oriented shading strokes. While none of the methods is perfect on this difficult instance,  the only ones to even partially pick up those patterns are our method and to some degree Gatys (but with strong color artifacts).
In general, we observe that oriented high-frequency patterns appear to benefit from higher (particularly, odd) moments, but further research is needed to explore the relation in depth.
\vspace{-1.5em}
\paragraph{Shape.}
Lastly, we turn our attention to shape. That attribute is somewhat more complex, as ornamental and decorative shape elements such as the square pattern in row 3 of Fig.~\ref{fig:user-study-results} are part of the style, whereas semantically meaningful elements of similar size are part of the content, like the eyes in row 4 or the make-up in row 5.
\ac{cmd} manages to disentangle these two aspects and preserve important boundaries and details of the content rather well, while still imposing the characteristic shape features of the style. Perhaps the most convincing example is row 3. But also in other cases the delicate balance between imposing the style and preserving salient content features appears to benefit from higher-order moments, \eg, rows 4, 5, 6.

\subsection{Quantitative results}

\paragraph{User study.}
There is no clear consensus how to quantitatively evaluate \ac{nst}. The question what constitutes a ``correct" output is clearly ill-posed, and even the judgment how ``good" a given stylization is depends on aesthetic preferences and must remain subjective.
In fact one can, with the \emph{same} method, generate very different results only by changing the relative weights of the style and content losses, and it depends on the application and on personal taste which one is preferred.

The current consensus is to perform user studies where participants are shown results without revealing how they were generated, and to collect statistics of user preferences. 
%
%
We note that, while we agree that aesthetic quality is hard to measure, people can usually pick their favorite among a handful of alternative stylizations without much hesitation, which lends some support to these studies: at the very least, they are a guideline which one among the available methods will deliver the result that the relatively largest share of the user group likes best.
We conduct a user study with the same methods as above: AdaIN \cite{huang2017arbitrary}, Gatys \cite{gatys2016image}, Moment Matching \cite{li2017demystifying}, OST \cite{lu2019closed}, WCT \cite{li2017universal} and the proposed \ac{cmd} method.
The study uses parts of the Kodak image dataset \cite{kodakdataset} and additional content images widely used in \ac{nst}, showing a variety of scenes, objects and humans.
%
The style dataset is made up by paintings and drawings commonly used for \ac{nst}, from a range of artists including Picasso, Kandinsky, Van Gogh and others.
In total we exhaustively combine 31 content images and 20 style images, resulting in 620 stylized images per algorithm. 
For the study, the six stylization results were displayed side-by-side in random order, along with the underlying content and style images. Users were asked to pick a single image that would best transfer style aspects such as shape, textures and colors using their own judgement.

Overall, we have collected $>$2700 votes from 56 different participants. The scores are reported in Tab.~\ref{tab:user-study}. The study reveals some interesting insights. Indeed, our proposed \ac{cmd} method performs favorably, with $\approx$10\% more votes than the closest competitor. The classical \ac{nst} of \cite{gatys2016image} attains the second-highest number of votes. This supports our claim that iterative methods still have an edge in terms of quality, as one-shot approaches trade quality for speed.

\begin{table}[!h]
    \setlength{\tabcolsep}{3pt}
    \renewcommand{\arraystretch}{1.4}
    \centering
    \begin{tabularx}{\linewidth}{|*6{>{\centering\arraybackslash}X}|}
    \hline
         AdaIN* & Gatys & MM & OST* & WCT* & Ours \\
         \hline
         155 & 533 & 443 & 523 & 463 & \textbf{587} \\
         5.7\% & 19.7\% & 16.3\% & 19.3\% & 17.1\% & \textbf{21.7\%} \\
         \hline
    \end{tabularx}
    \caption{Number of votes each method received in our user study. * denotes one-shot feed-forward methods.}
    \label{tab:user-study}
\end{table}


\makeatletter
\newlength{\tw}
\setlength{\tw}{0.095\textwidth}
\makeatother

\begin{figure}[!tb]
\centering
\setlength{\tabcolsep}{0pt}
\renewcommand{\arraystretch}{0.1}
\begin{tabular}{ccccc}

\vspace{2pt}
\footnotesize 1\textsuperscript{st} moment &
\footnotesize 2\textsuperscript{nd} moment &
\footnotesize 3\textsuperscript{rd} moment &
\footnotesize 4\textsuperscript{th} moment &
\footnotesize 5\textsuperscript{th} moment
\\
\includegraphics[width=\tw]{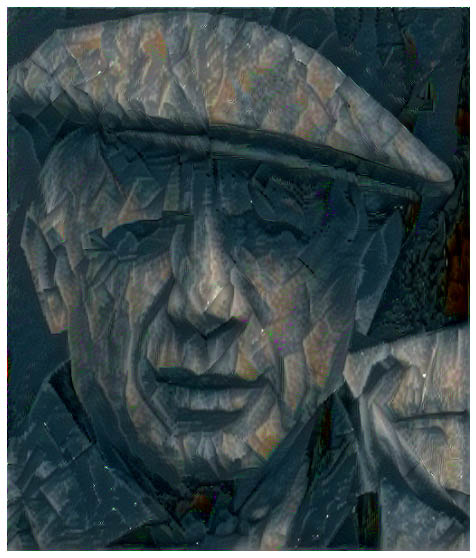} &
\includegraphics[width=\tw]{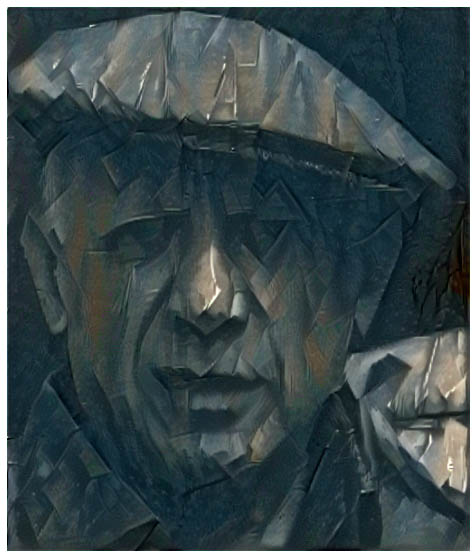} &
\includegraphics[width=\tw]{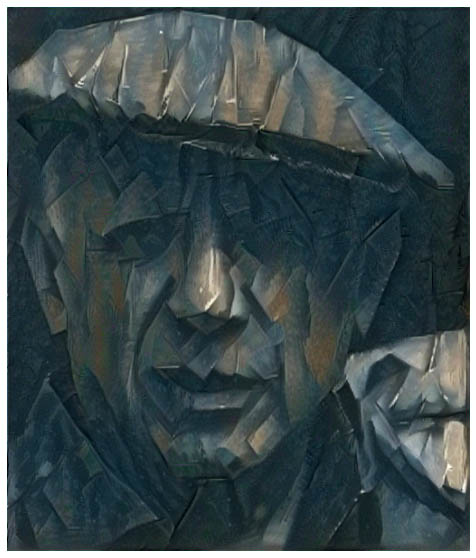} &
\includegraphics[width=\tw]{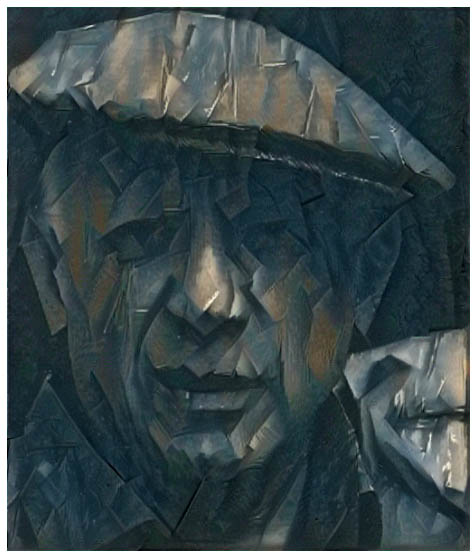} &
\includegraphics[width=\tw]{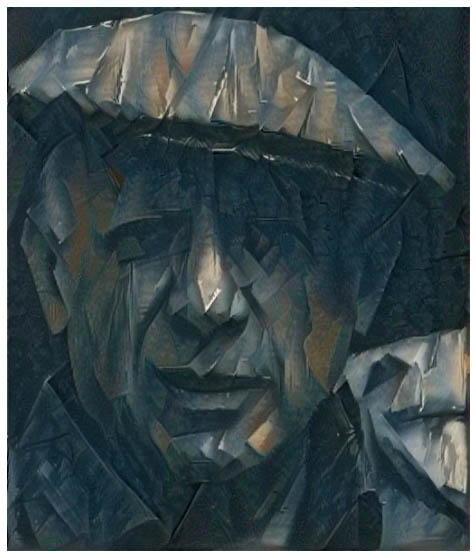} \\
&
\includegraphics[width=\tw]{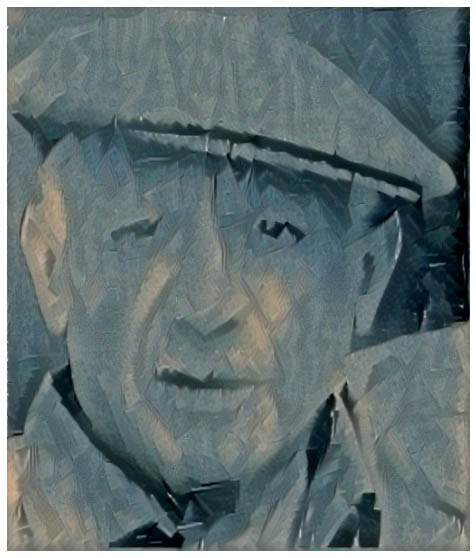} &
\includegraphics[width=\tw]{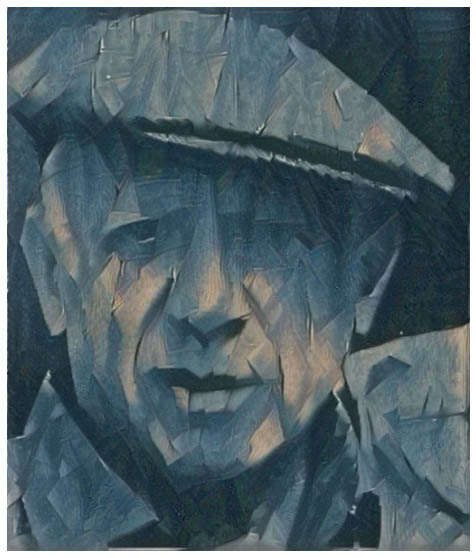} &
\includegraphics[width=\tw]{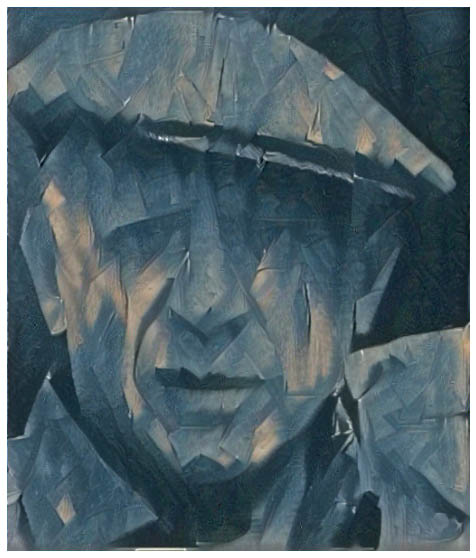} &
\includegraphics[width=\tw]{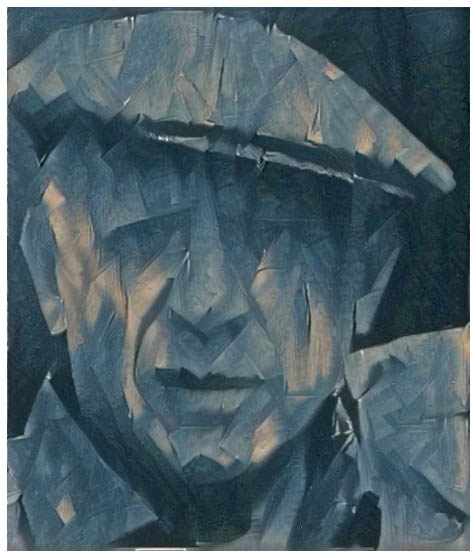} \\
\raisebox{3pt}{\footnotesize Content} &&
\includegraphics[width=\tw]{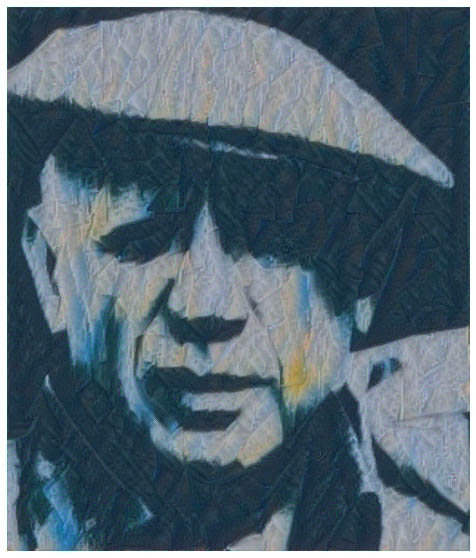} &
\includegraphics[width=\tw]{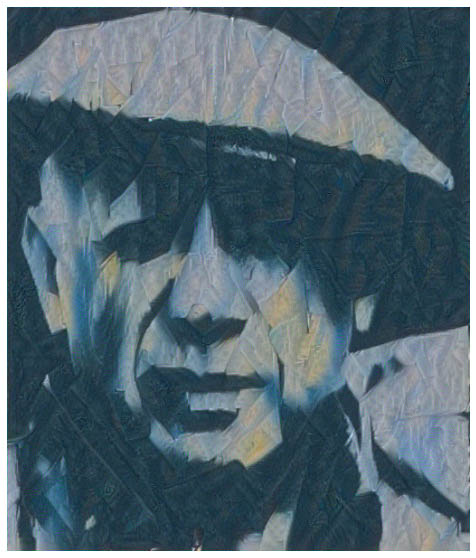} &
\includegraphics[width=\tw]{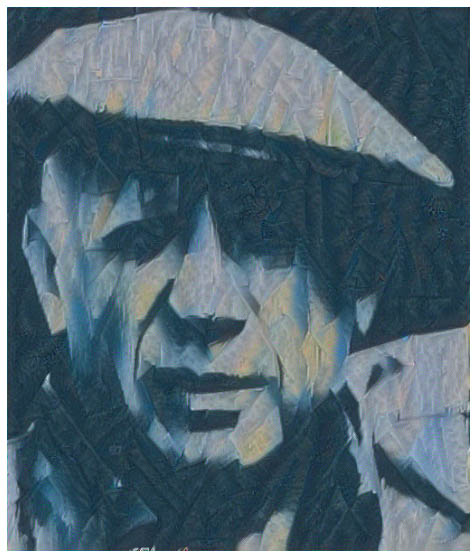} \\
\includegraphics[width=\tw]{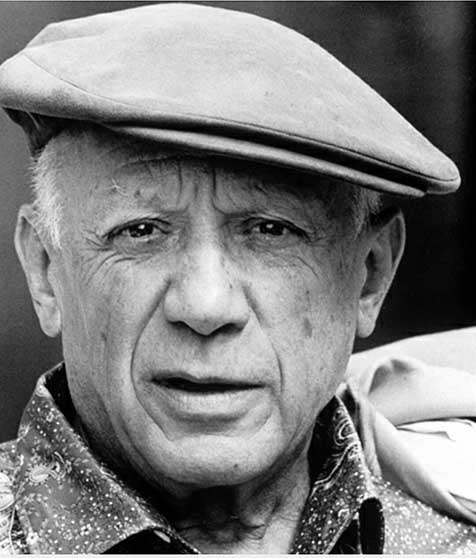} &
\raisebox{3pt}{\footnotesize Style} & &
\includegraphics[width=\tw]{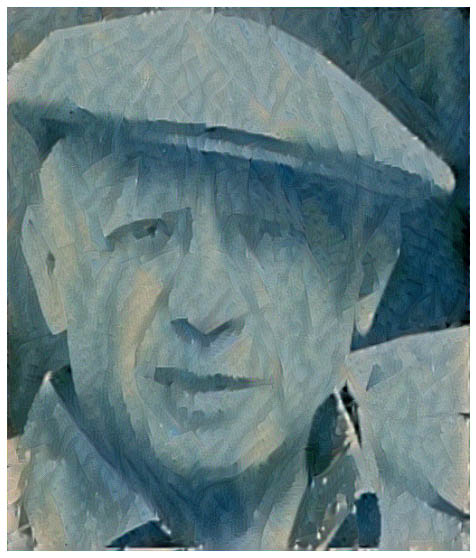} &
\includegraphics[width=\tw]{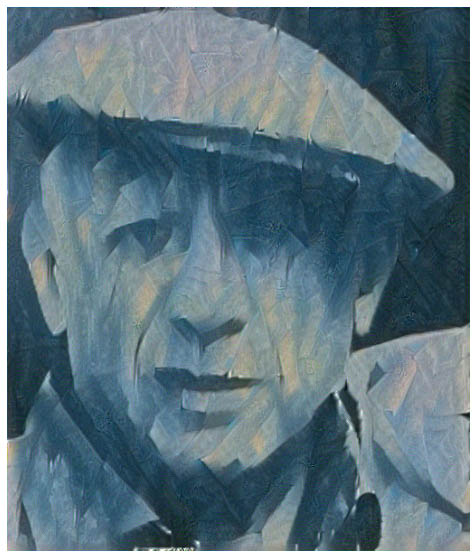} \\ 
&
\includegraphics[width=\tw]{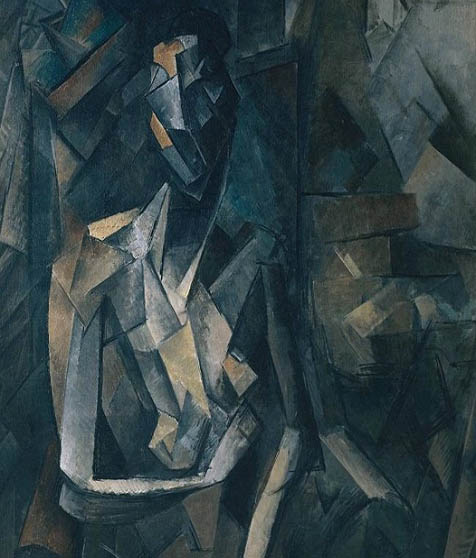} &
&&
\includegraphics[width=\tw]{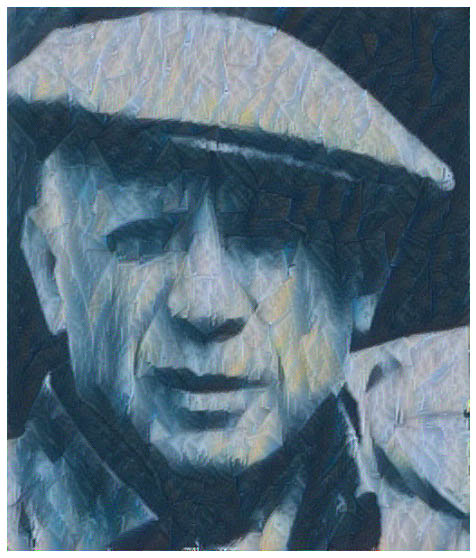}
\end{tabular}
\caption{Ablation study using only selected moments. See text for details.}\label{fig:momentablation}
\end{figure}
\vspace{-0.5em}
\subsection{Ablation studies}\label{sec:ablationstudies}

In our method it is possible to individually reweight or turn off moments. We have conducted an ablation study to better understand the effects of different moments, see Fig.~\ref{fig:momentablation}.
Note that this tuning knob is orthogonal to user control in the spirit of \cite{gatys2017controlling}, where one isolates a specific attribute like color in preprocessing and applies the stylization selectively. 
Figure \ref{fig:momentablation} shows style transfer results with different combinations of moments. Only a single moment corresponding to the row/column index is used on the diagonal. Then higher-order moments are progressively added along the rows, so for instance position $(2,2)$ corresponds to only the second moment (weight vector $a=[0,1,0,0,0]$) and element $(2,4)$ corresponds to the the 2\textsuperscript{nd}, 3\textsuperscript{rd} and 4\textsuperscript{th} moments (weight vector $a=[0,1,1,1,0]$).
As was to be expected there is no obvious, ``pure" correspondence between moments and visual attributes. Still, the study illustrates some interesting relations. First, one can immediately see that even the 5\textsuperscript{th} order still contributes significant style elements, for instance on the chin and the cap in the first row.
Odd moments appear to primarily modulate overall brightness and contrast, whereas even ones tend to change colors and high-frequency texture.

\begin{figure}[!tb]
    \centering
    \begin{subfigure}[b]{0.32\linewidth}
        \centering
        \includegraphics[width=\textwidth]{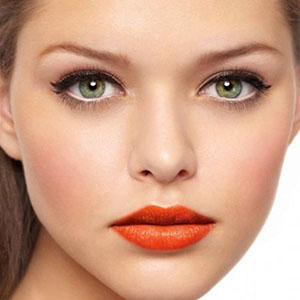}
        \caption{Content}
    \end{subfigure}
    \begin{subfigure}[b]{0.32\linewidth}
        \centering
        \includegraphics[width=\textwidth]{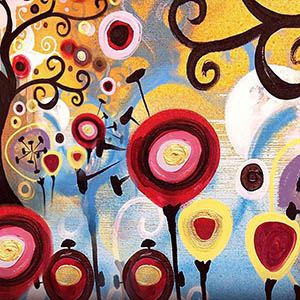}
        \caption{Style}
    \end{subfigure}
    \begin{subfigure}[b]{0.32\linewidth}
        \centering
        \includegraphics[width=\textwidth]{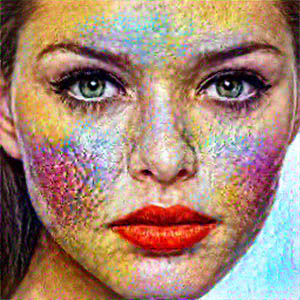}
        \caption{$\alpha=0.6$}
    \end{subfigure}
    \par\medskip
    \begin{subfigure}[b]{0.32\linewidth}
        \centering
        \includegraphics[width=\textwidth]{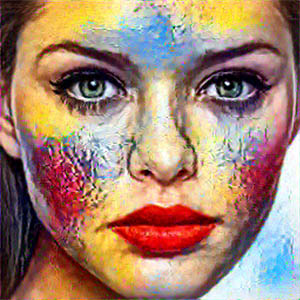}
        \caption{$\alpha=0.2$}
    \end{subfigure}
    \begin{subfigure}[b]{0.32\linewidth}
        \centering
        \includegraphics[width=\textwidth]{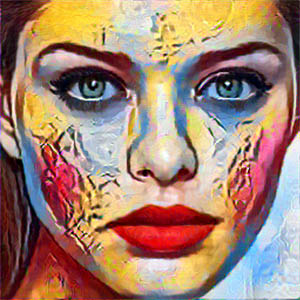}
        \caption{$\alpha=0.01$}
    \end{subfigure}
    \begin{subfigure}[b]{0.32\linewidth}
        \centering
        \includegraphics[width=\textwidth]{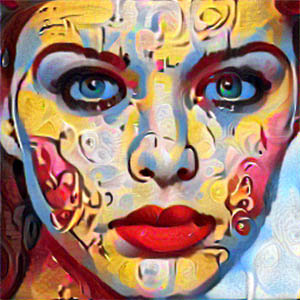}
        \caption{$\alpha=0$}
    \end{subfigure}
    \caption{Varying the strength of style transfer by varying the relative influence $\alpha$ of the content loss (\cf eq.~\eqref{eq:nstloss}).}
    \label{fig:linearinterpolation}
\end{figure}

Our \ac{cmd} method changes only the loss function for distribution alignment and can be seamlessly combined with 
other extensions of \ac{nst}. For instance, the user can still control how strongly the style is imprinted on the image content, by adjusting the relative weight of the style and content losses.
To illustrate this, we stylize with our \ac{cmd} method and linearly interpolate the weight $\alpha$ in eq.~\eqref{eq:nstloss}.
Figure \ref{fig:linearinterpolation} shows an example how putting more weight on the content loss produces increasingly weaker "partial stylizations" that stay closer to the content image.
%

\section{Limitations and future work}
There are currently two conceptual directions in \ac{nst}: iterative optimization techniques and one-shot feed-forward approaches. Our algorithm belongs to the former.
While iterative methods arguably still produce better results, they are too slow for real-time applications. Our method inherits that shortcoming, \eg, it could not be used for (near) real-time video synthesis.

At the conceptual level, we had to make two simplifying approximations to take the step from the mathematical formalism of \ac{cmd} to a practical implementation.
On the one hand, we limit the order of the central moments to a finite, in practice small $K$.
At least in principle the impact of that restriction can be kept as small as desired by increasing $K$, because the influence of additional central moments provably converges $\rightarrow\!0$ with increasing order.

On the other hand, and perhaps more importantly, we only utilize the \textit{marginal} central moments in our loss.
We take this shortcut for computational reasons, but it effectively means that we only achieve exact distribution matching when the marginal distributions are independent. 
There is currently no evidence that this is the case, and we do not see a simple way to gauge how much information might be lost due to the approximation.

\section{Conclusion}
We have revisited the interpretation of \acl{nst} as aligning feature distributions. After categorizing existing methods into three groups based on \ac{mmd}, moment matching and \ac{ot}, we show that all of them, in practice, only match first and second moments.
We then went on to propose a novel approach based on \aclp{cmd}. Our method can be interpreted alternatively as minimizing an integral probability metric, or as matching all central moments up to a desired order.
Our method has both theoretical and practical benefits. In terms of theory it comes with strong approximation guarantees. On the practical side it offers a computationally efficient way to account for higher-order moments of complex feature distributions, and achieves visually better transfer of many artistic styles.
On a broader scale, even though Portilla and Simoncelli proposed higher order matching to texture synthesis \cite{portilla2000parametric}, Gatys \etal \cite{gatys2015texture, gatys2016image} disregarded all but second-order moments when pioneering Neural Style Transfer. In this regard, our method reintroduces higher order matching to \ac{nst}.

{\small
\bibliographystyle{ieee_fullname}
\bibliography{bibliography}
}

\onecolumn
\newpage

\appendix
\appendixpage
\addappheadtotoc

\section{Run time experiments}

We document the run time of our proposed algorithm and compare it to that of our baselines, using the implementations provided by the authors (\cf footnote 1 in the main paper).
The timings are for images of size 512$\times$512 pixels, and running all iterative methods for 500 iterations. We average over 15 runs on a single Nvidia GeForce GTX 1080Ti. The results are shown in Tab.~\ref{tab:runtime-study}.
Naturally, one-shot feed-forward methods are a lot faster to compute, at the cost of a bit lower image quality.
Among the iterative methods, the differences are practically negligible. Ours is on par with the two competitors, adding \textless10\% of computational overhead over Gatys' original method; while being slightly faster than MM, due to a more efficient implementation.

\begin{table}[!h]
    \setlength{\tabcolsep}{3pt}
    \renewcommand{\arraystretch}{1.4}
    \centering
    \vspace{0.5em}
    \begin{tabularx}{\linewidth}{|*6{>{\centering\arraybackslash}X}|}
    \hline
         AdaIN* & Gatys & MM & OST* & WCT* & Ours \\
         \hline
         0.58s & 30.51s & 35.49s & 2.40s & 1.93s & \textbf{33.59s} \\
         \hline
    \end{tabularx}
    \caption{Run time of different \ac{nst} methods, in seconds.}
    \label{tab:runtime-study}
\end{table}

\section{Influence of the learning rate}

\begin{figure}[!htbp]
    \centering
    \begin{subfigure}[b]{0.16\linewidth}
        \centering
        \includegraphics[width=\textwidth]{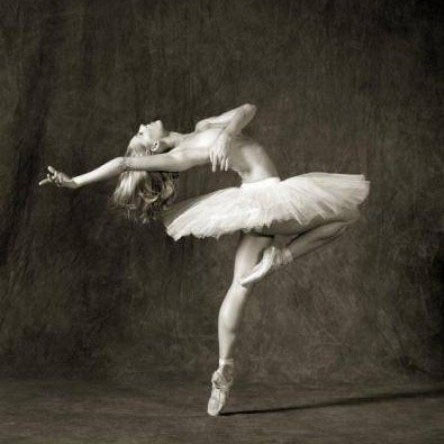}
        \caption{Content}
    \end{subfigure}
    \begin{subfigure}[b]{0.16\linewidth}
        \centering
        \includegraphics[width=\textwidth]{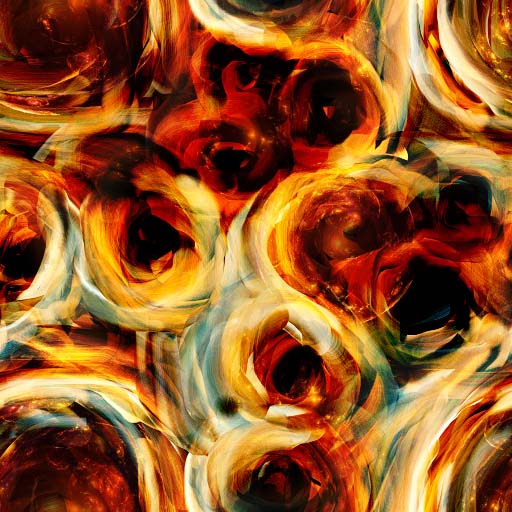}
        \caption{Style}
    \end{subfigure}
    \begin{subfigure}[b]{0.16\linewidth}
        \centering
        \includegraphics[width=\textwidth]{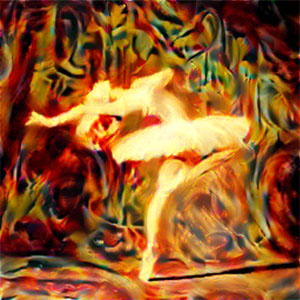}
        \caption{$\text{lr}=0.01$}
    \end{subfigure}
    \begin{subfigure}[b]{0.16\linewidth}
        \centering
        \includegraphics[width=\textwidth]{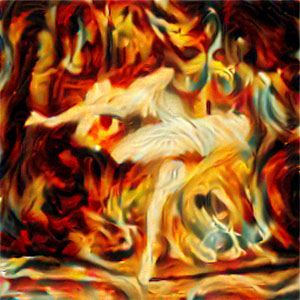}
        \caption{$\text{lr}=0.1$}
    \end{subfigure}
    \begin{subfigure}[b]{0.16\linewidth}
        \centering
        \includegraphics[width=\textwidth]{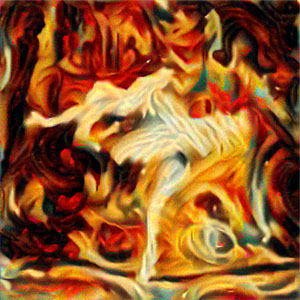}
        \caption{$\text{lr}=0.2$}
    \end{subfigure}
    \begin{subfigure}[b]{0.16\linewidth}
        \centering
        \includegraphics[width=\textwidth]{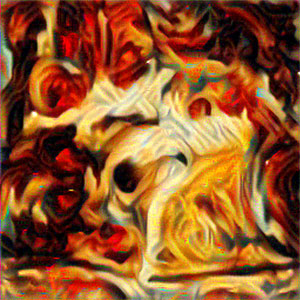}
        \caption{$\text{lr}=0.3$}
    \end{subfigure}
    \begin{subfigure}[b]{0.16\linewidth}
        \centering
        \includegraphics[width=\textwidth]{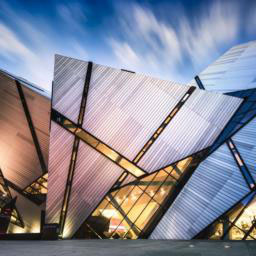}
        \caption{Content}
    \end{subfigure}
    \begin{subfigure}[b]{0.16\linewidth}
        \centering
        \includegraphics[width=\textwidth]{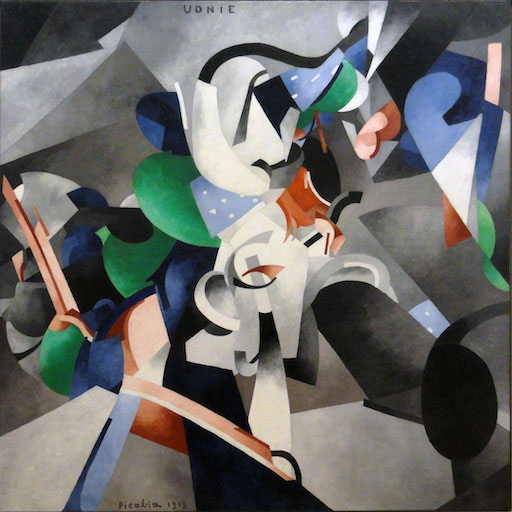}
        \caption{Style}
    \end{subfigure}
    \begin{subfigure}[b]{0.16\linewidth}
        \centering
        \includegraphics[width=\textwidth]{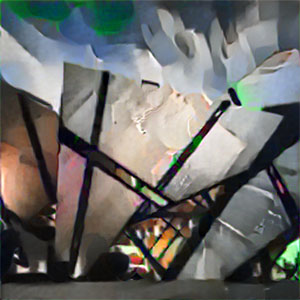}
        \caption{$\text{lr}=0.01$}
    \end{subfigure}
    \begin{subfigure}[b]{0.16\linewidth}
        \centering
        \includegraphics[width=\textwidth]{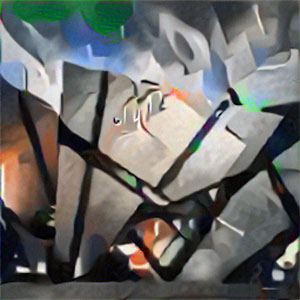}
        \caption{$\text{lr}=0.1$}
    \end{subfigure}
    \begin{subfigure}[b]{0.16\linewidth}
        \centering
        \includegraphics[width=\textwidth]{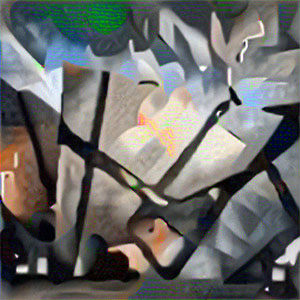}
        \caption{$\text{lr}=0.2$}
    \end{subfigure}
    \begin{subfigure}[b]{0.16\linewidth}
        \centering
        \includegraphics[width=\textwidth]{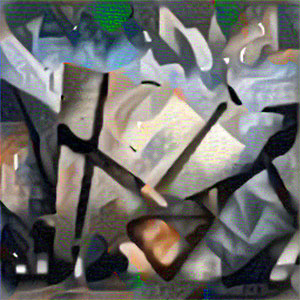}
        \caption{$\text{lr}=0.3$}
    \end{subfigure}
    \caption{Impact of learning rate on the output of our \ac{cmd} method.}
    \label{fig:differentlr}
\end{figure}

We further investigate the influence of varying learning rates. As can be seen from Fig.~\ref{fig:differentlr}, increasing the learning rate has a similar effect as reducing the weight $\alpha$ of the content loss in \eqref{eq:nstloss}. This is expected, as the style loss can be decreased more rapidly when disregarding the "constraint" to preserve the content, encoded in the content loss. %
With too high learning rate, only barely recognisable traces of the image content are preserved, as can be seen towards the right side of Fig.~\ref{fig:differentlr}. Also, training becomes increasingly unstable, as often for deep networks one must balance learning speed against learning success.

\newpage
\section{Additional qualitative results}
\vspace{-0.6em}
\setlength{\tabcolsep}{1pt}
\renewcommand{\arraystretch}{0.2}

\begin{figure}[!htbp]
    \begin{center}
        \begin{tabularx}{\textwidth}{*{8}{>{\centering\arraybackslash}X}}
              \includegraphics[width=0.124\textwidth]{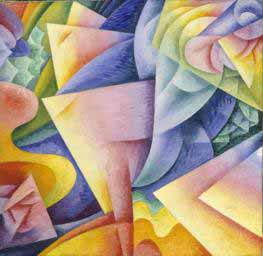}
              & \includegraphics[width=0.124\textwidth]{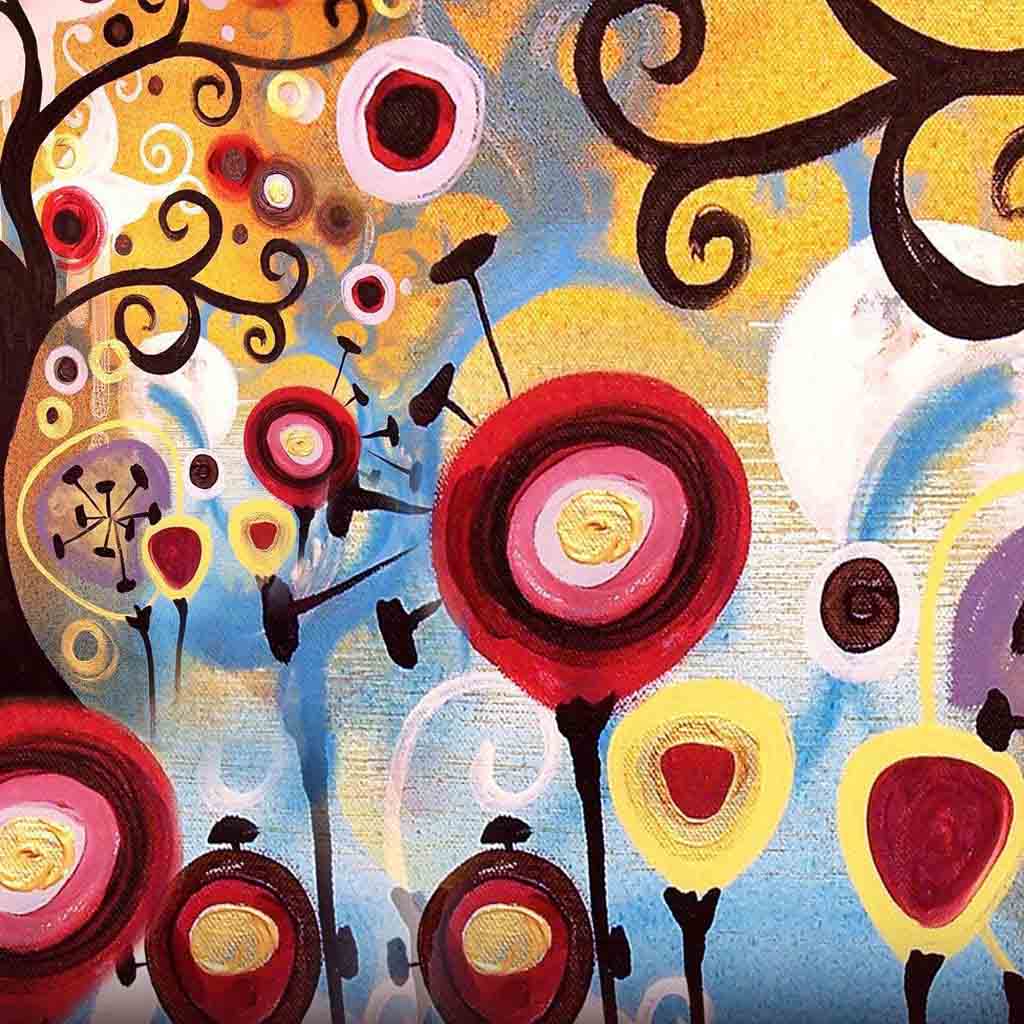}
              & \includegraphics[width=0.124\textwidth]{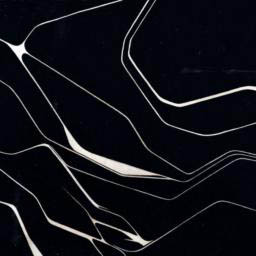}
              & \includegraphics[width=0.124\textwidth]{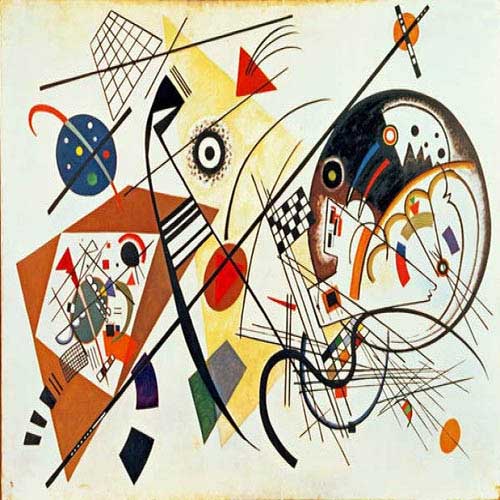}
              & \includegraphics[width=0.124\textwidth]{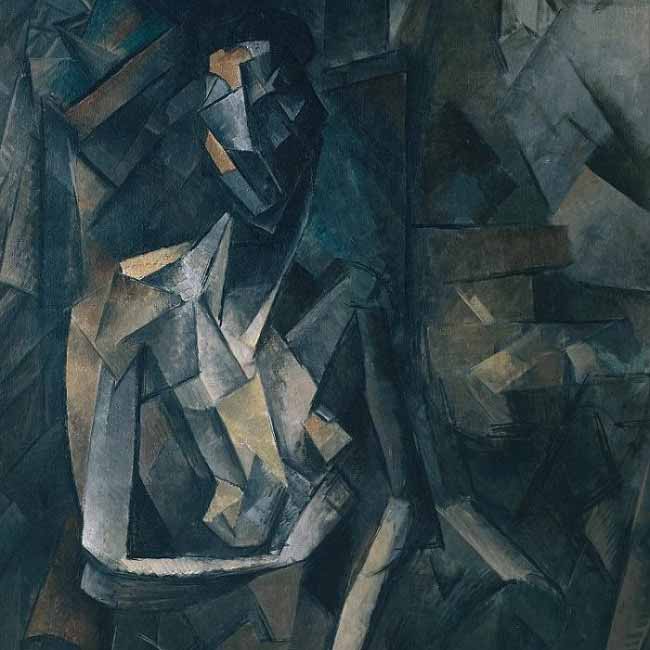}
              & \includegraphics[width=0.124\textwidth]{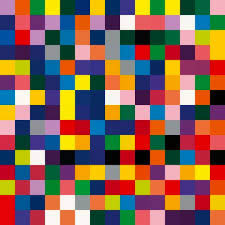}
              & \includegraphics[width=0.124\textwidth]{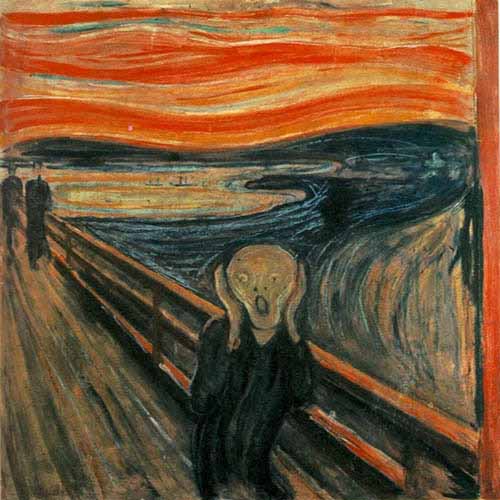}
              & \includegraphics[width=0.124\textwidth]{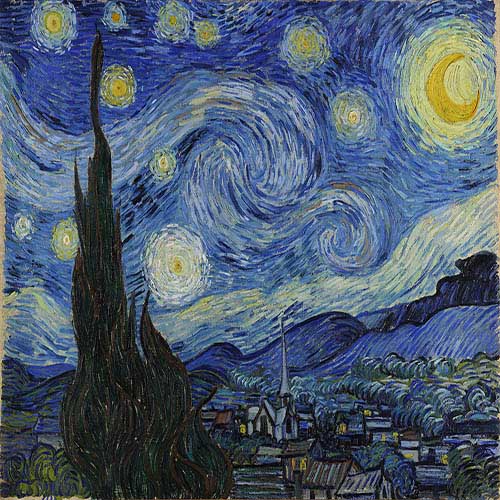}\\
              \includegraphics[width=0.124\textwidth]{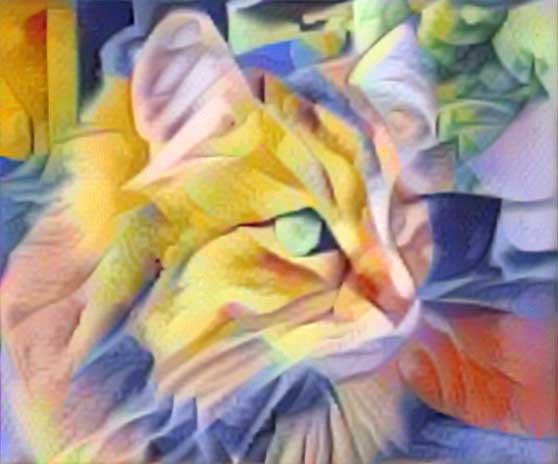}
              & \includegraphics[width=0.124\textwidth]{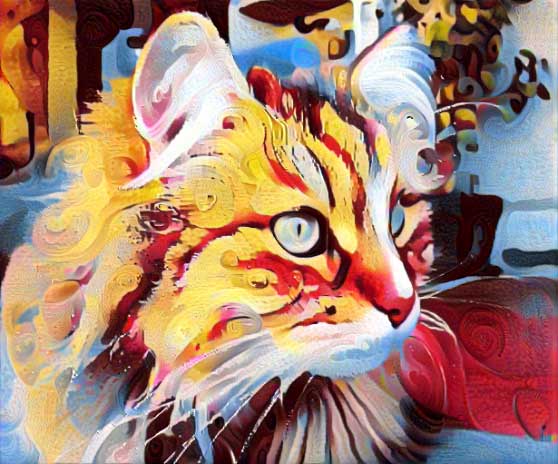}
              & \includegraphics[width=0.124\textwidth]{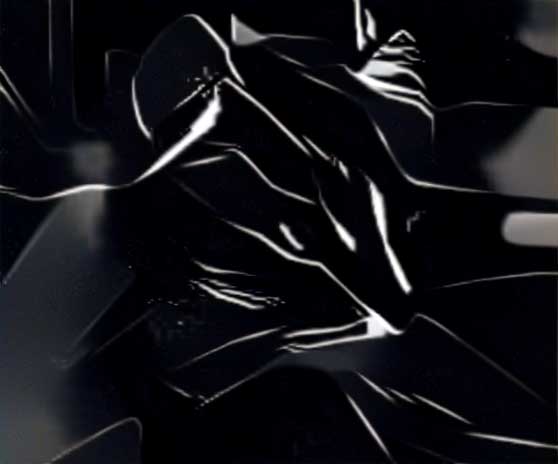}
              & \includegraphics[width=0.124\textwidth]{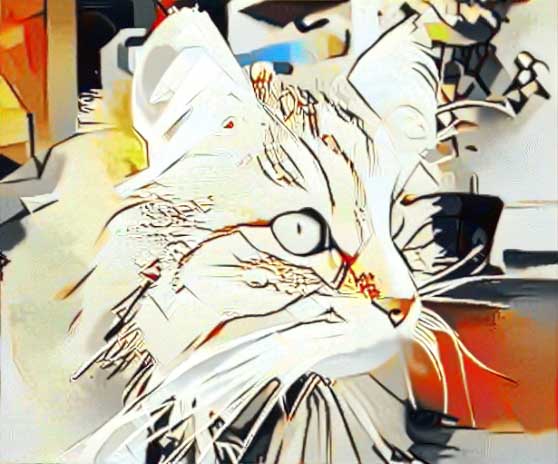}
              & \includegraphics[width=0.124\textwidth]{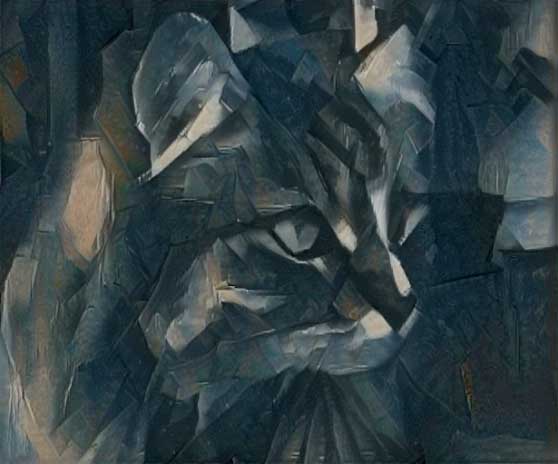}
              & \includegraphics[width=0.124\textwidth]{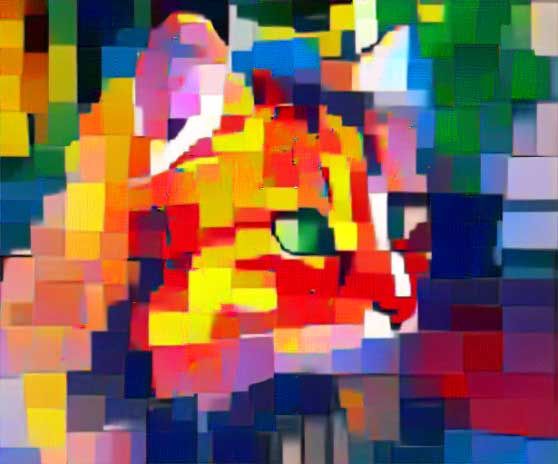}
              & \includegraphics[width=0.124\textwidth]{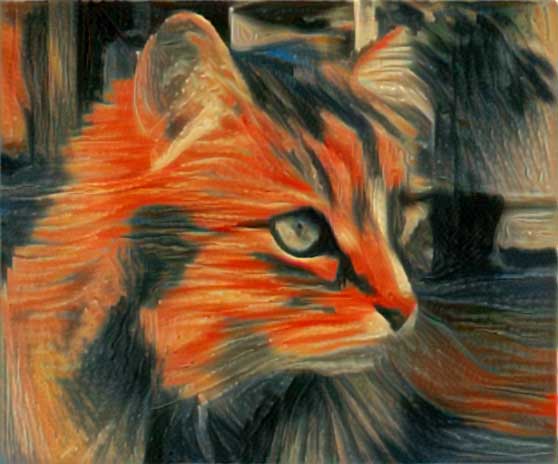}
              & \includegraphics[width=0.124\textwidth]{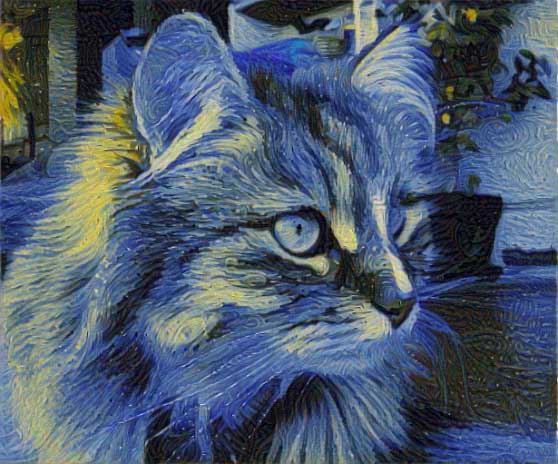}\\
              \includegraphics[width=0.124\textwidth]{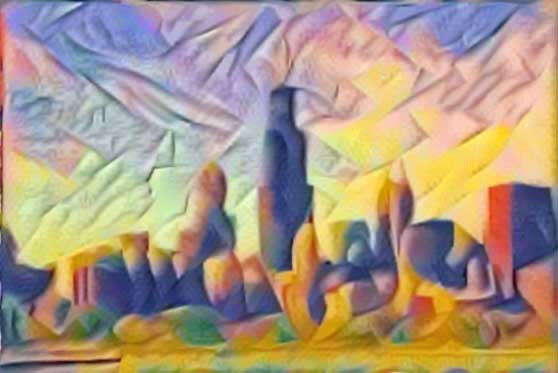}
              & \includegraphics[width=0.124\textwidth]{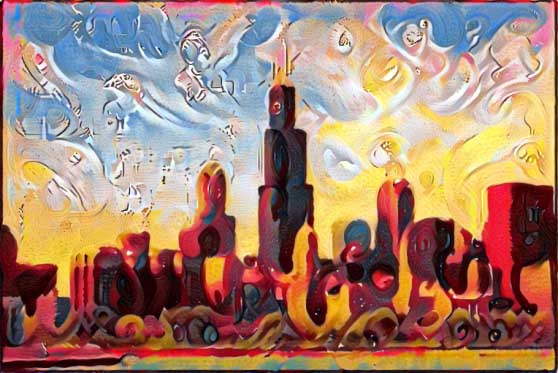}
              & \includegraphics[width=0.124\textwidth]{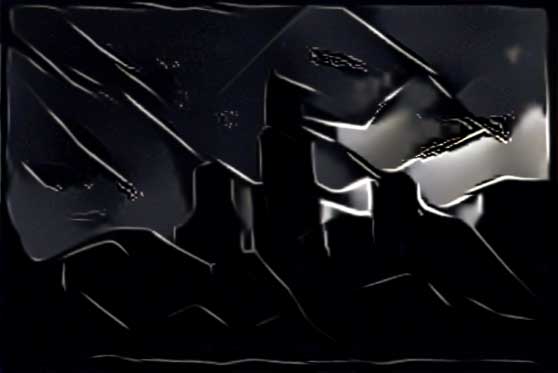}
              & \includegraphics[width=0.124\textwidth]{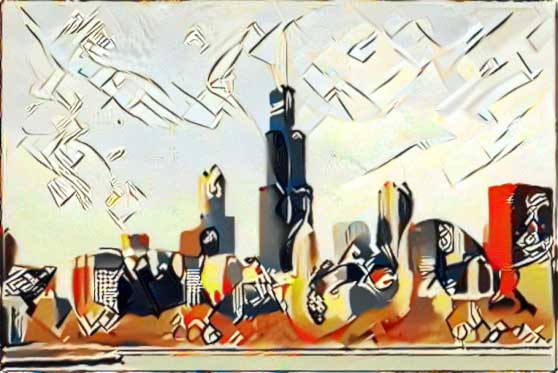}
              & \includegraphics[width=0.124\textwidth]{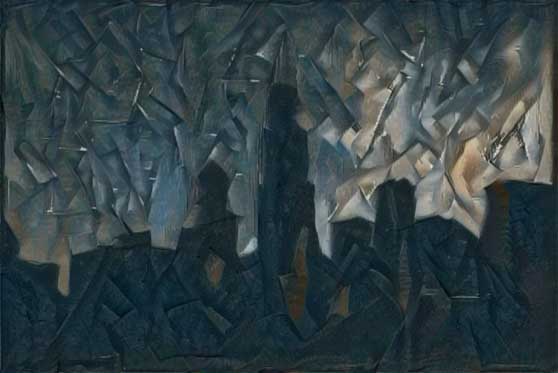}
              & \includegraphics[width=0.124\textwidth]{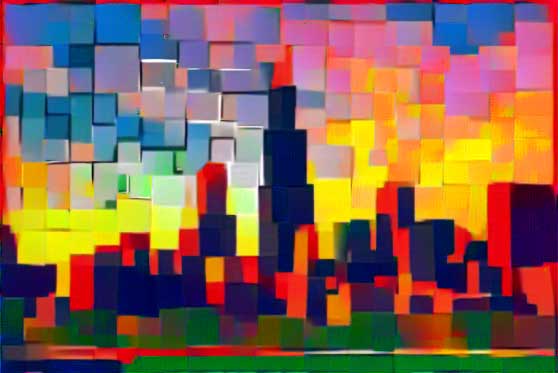}
              & \includegraphics[width=0.124\textwidth]{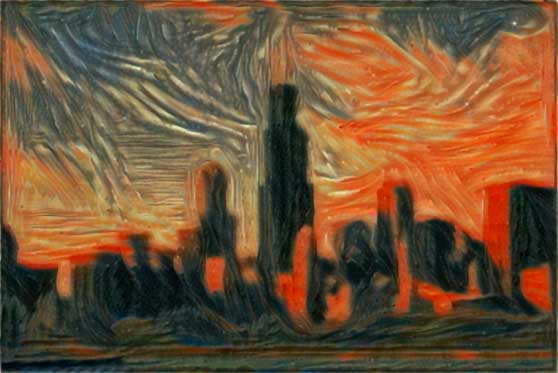}
              & \includegraphics[width=0.124\textwidth]{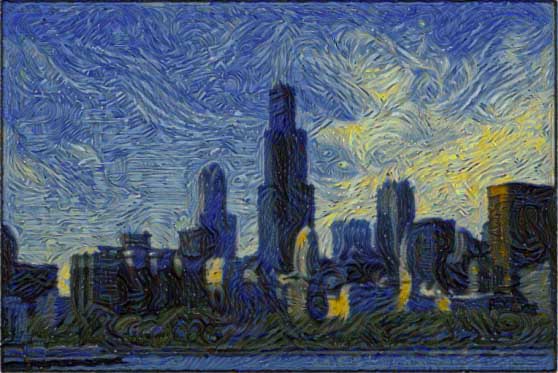}\\
              \includegraphics[width=0.124\textwidth]{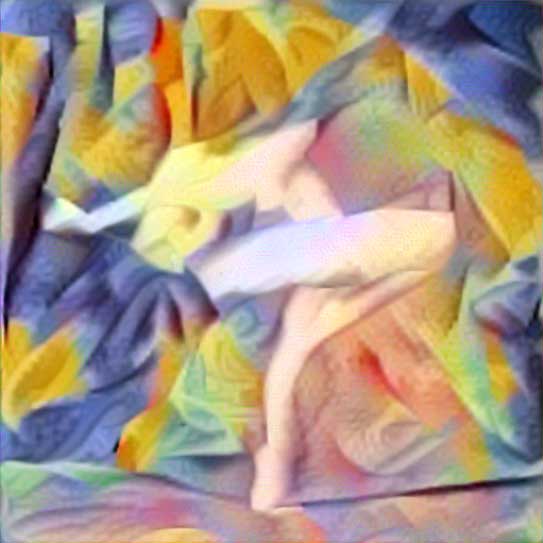}
              & \includegraphics[width=0.124\textwidth]{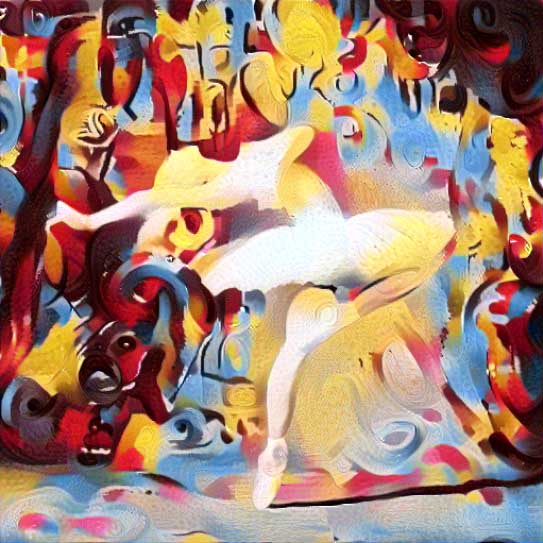}
              & \includegraphics[width=0.124\textwidth]{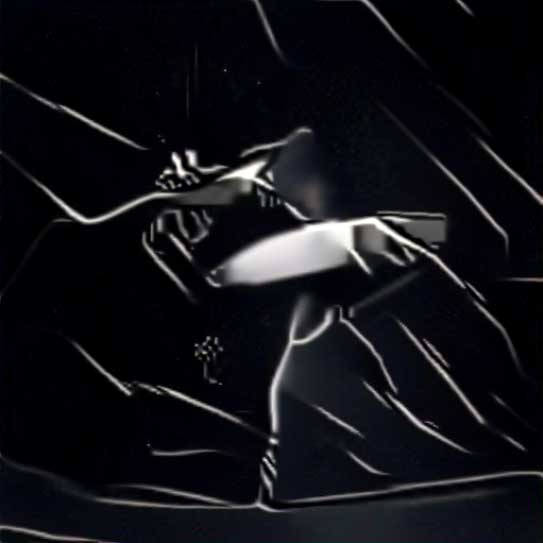}
              & \includegraphics[width=0.124\textwidth]{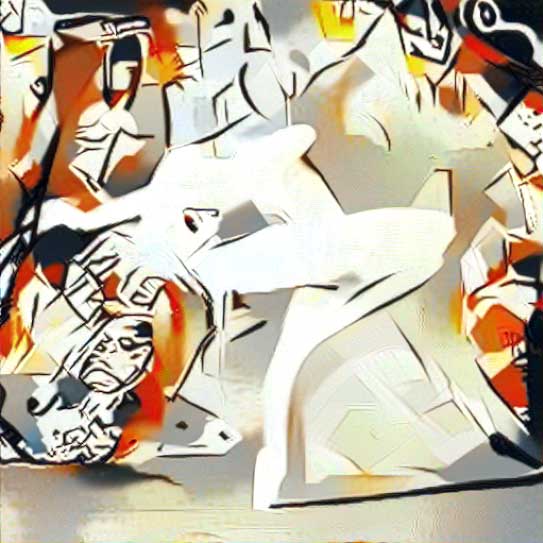}
              & \includegraphics[width=0.124\textwidth]{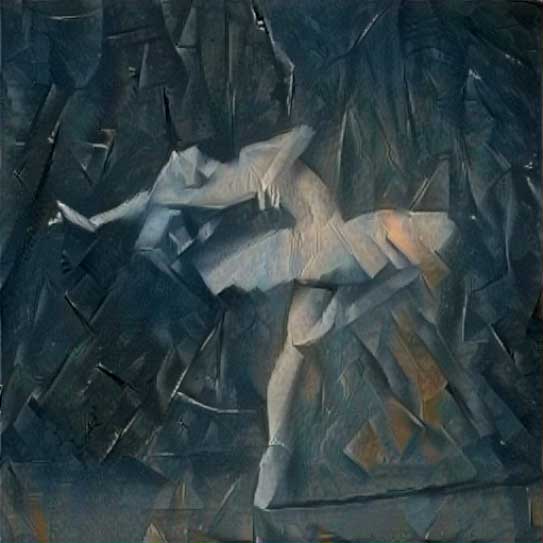}
              & \includegraphics[width=0.124\textwidth]{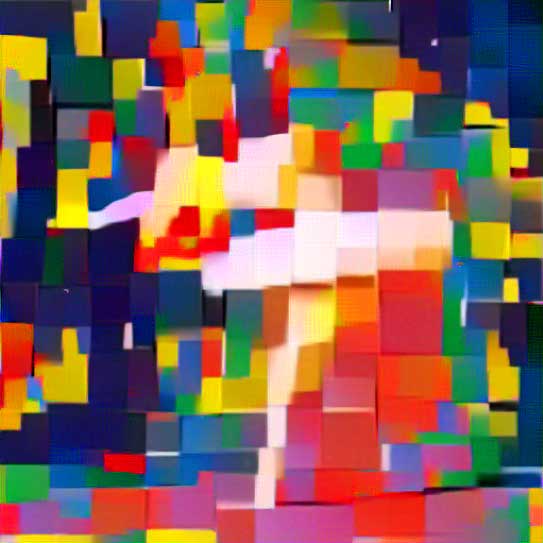}
              & \includegraphics[width=0.124\textwidth]{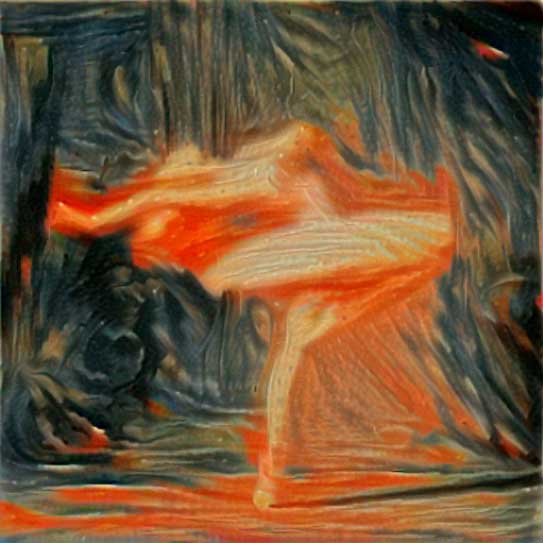}
              & \includegraphics[width=0.124\textwidth]{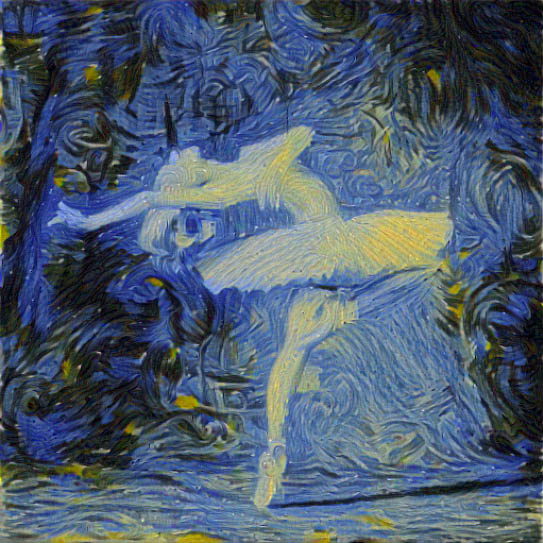}\\
              \includegraphics[width=0.124\textwidth]{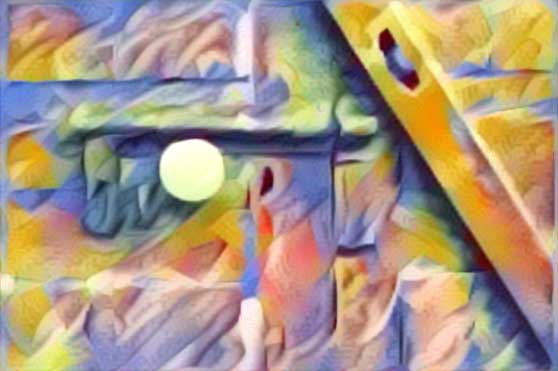}
              & \includegraphics[width=0.124\textwidth]{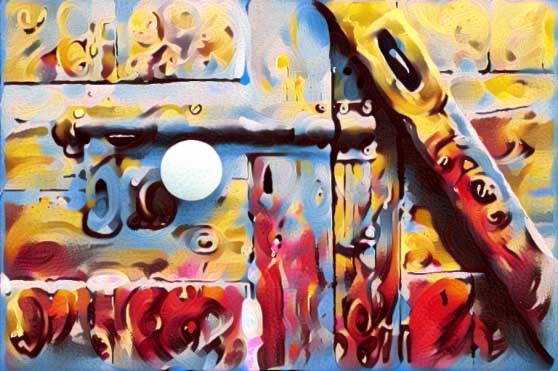}
              & \includegraphics[width=0.124\textwidth]{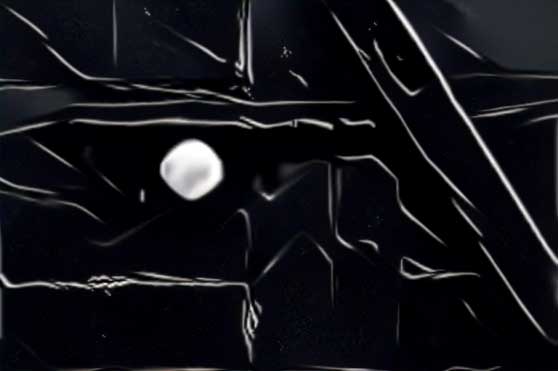}
              & \includegraphics[width=0.124\textwidth]{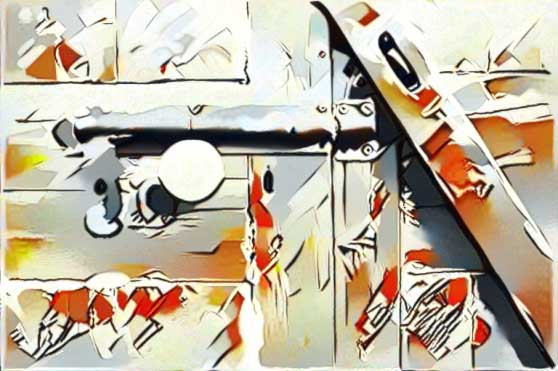}
              & \includegraphics[width=0.124\textwidth]{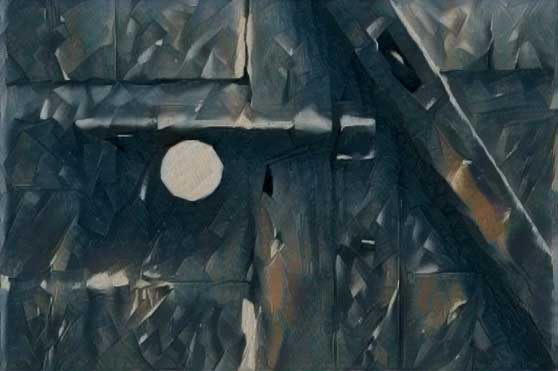}
              & \includegraphics[width=0.124\textwidth]{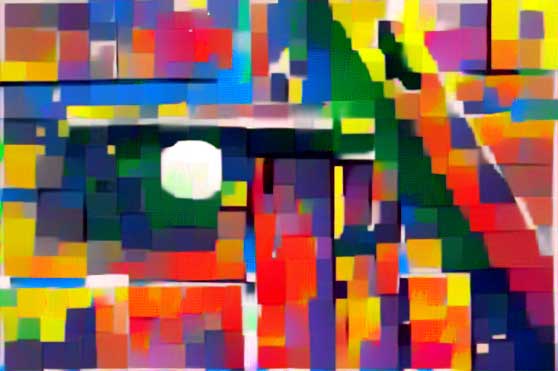}
              & \includegraphics[width=0.124\textwidth]{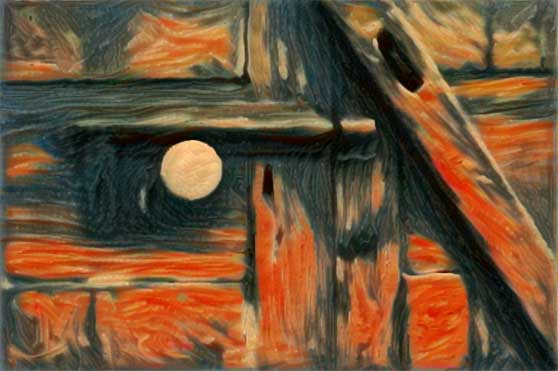}
              & \includegraphics[width=0.124\textwidth]{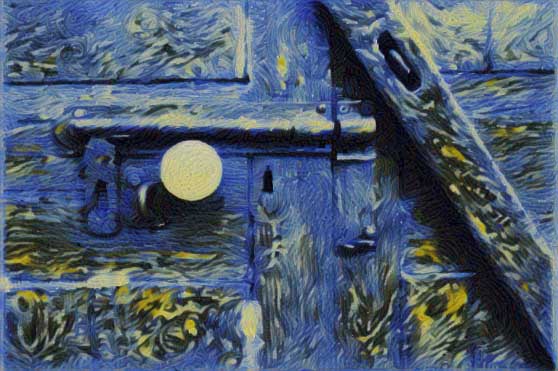}\\
              \includegraphics[width=0.124\textwidth]{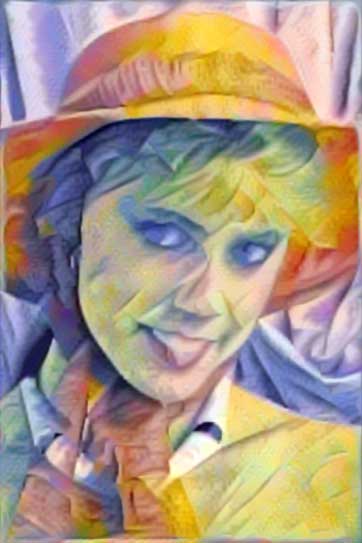}
              & \includegraphics[width=0.124\textwidth]{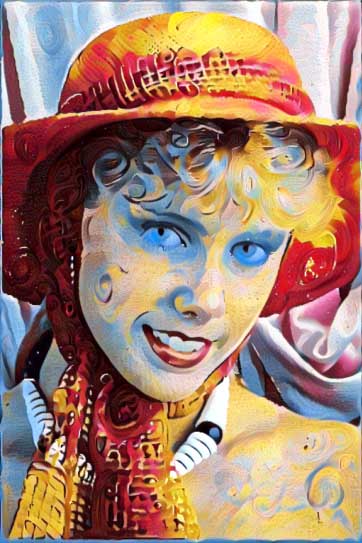}
              & \includegraphics[width=0.124\textwidth]{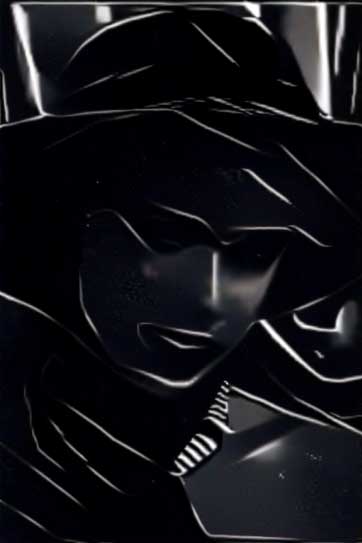}
              & \includegraphics[width=0.124\textwidth]{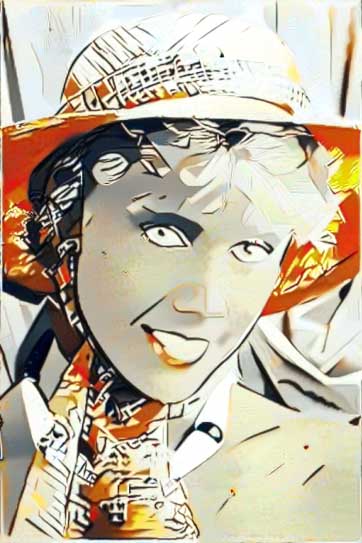}
              & \includegraphics[width=0.124\textwidth]{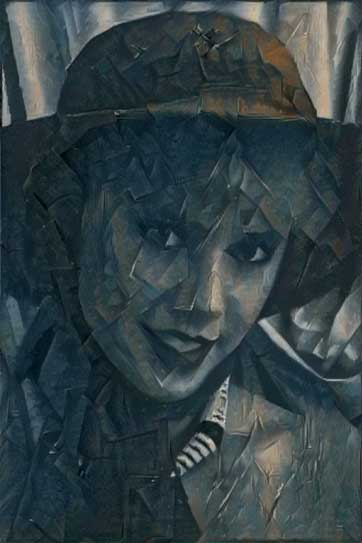}
              & \includegraphics[width=0.124\textwidth]{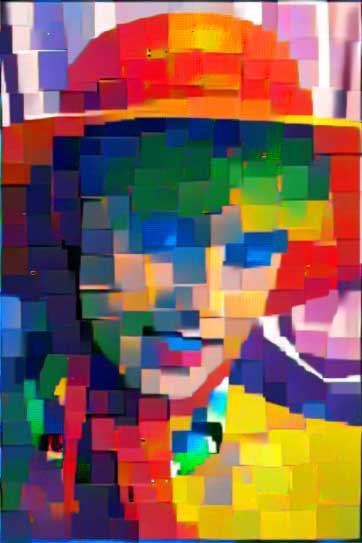}
              & \includegraphics[width=0.124\textwidth]{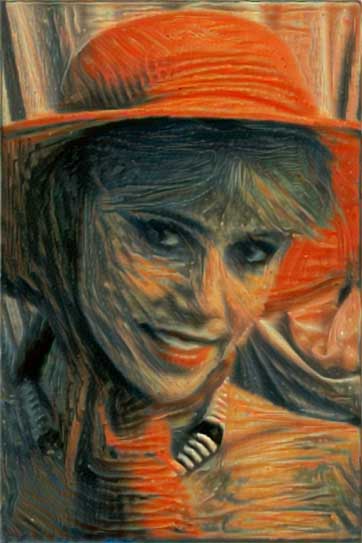}
              & \includegraphics[width=0.124\textwidth]{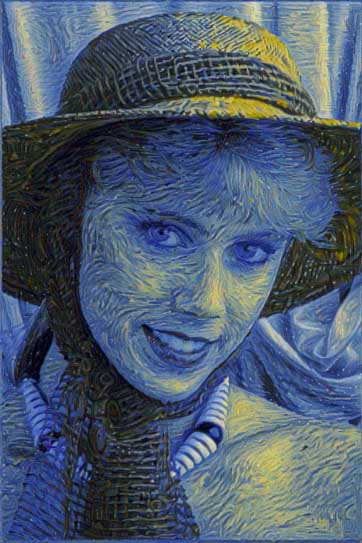}\\
              \includegraphics[width=0.124\textwidth]{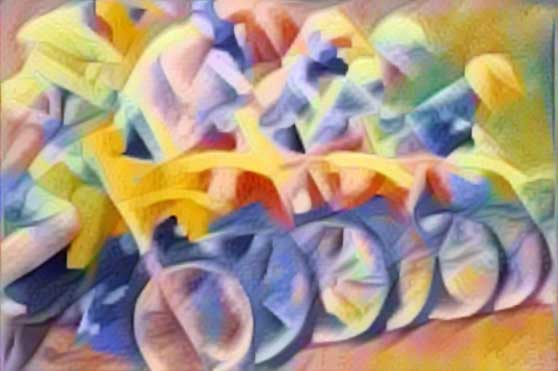}
              & \includegraphics[width=0.124\textwidth]{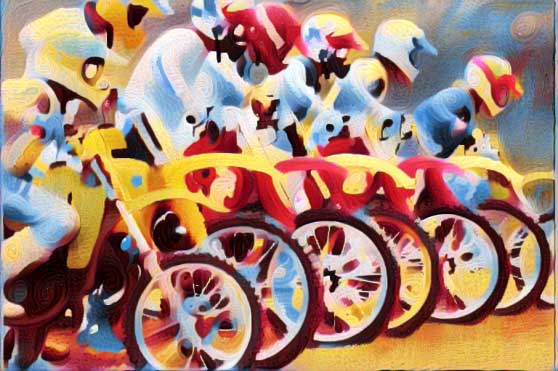}
              & \includegraphics[width=0.124\textwidth]{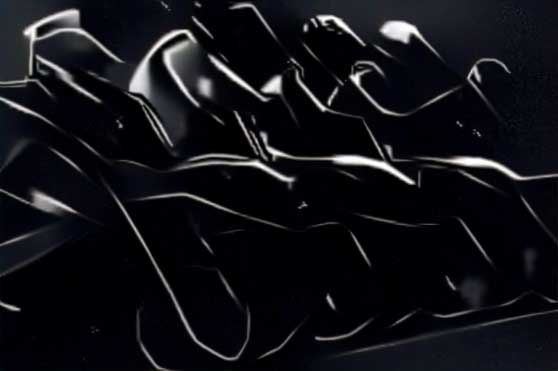}
              & \includegraphics[width=0.124\textwidth]{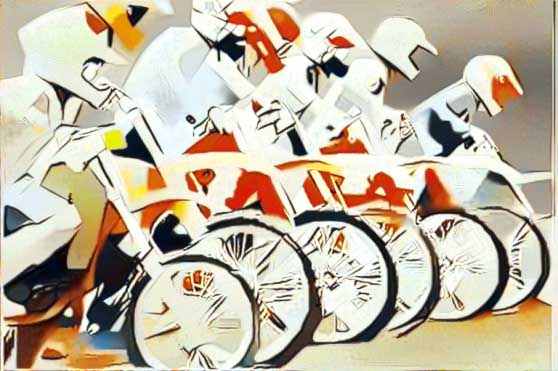}
              & \includegraphics[width=0.124\textwidth]{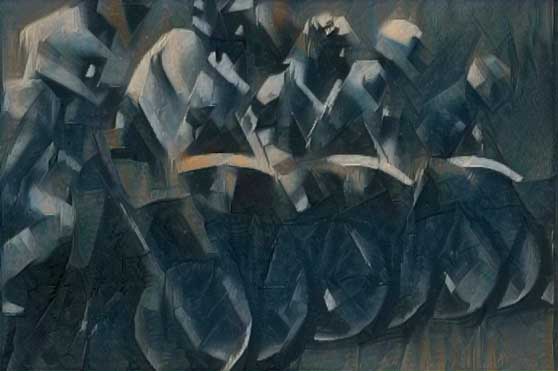}
              & \includegraphics[width=0.124\textwidth]{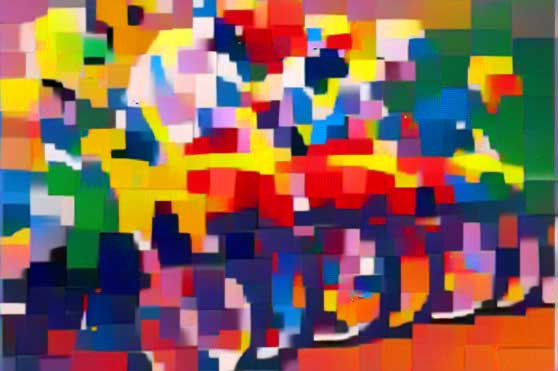}
              & \includegraphics[width=0.124\textwidth]{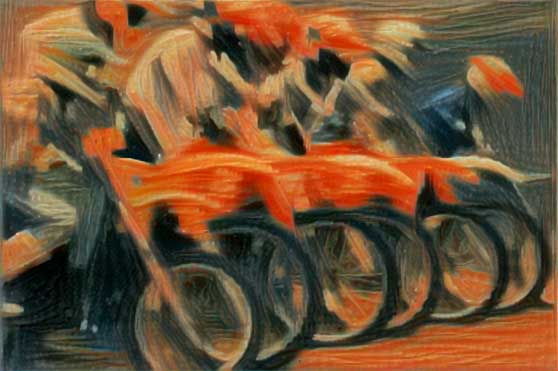}
              & \includegraphics[width=0.124\textwidth]{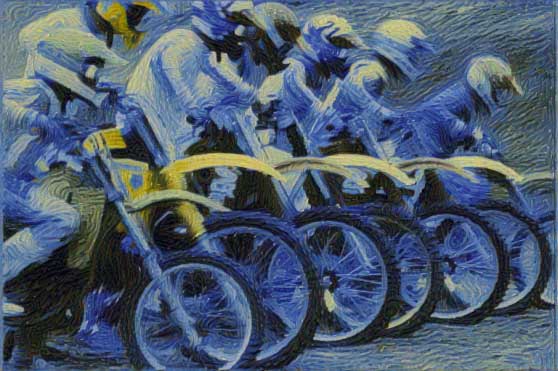}\\
              \includegraphics[width=0.124\textwidth]{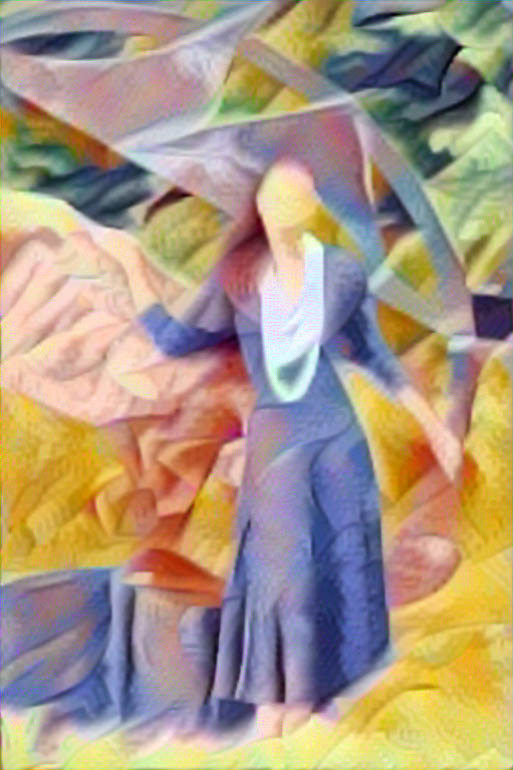}
              & \includegraphics[width=0.124\textwidth]{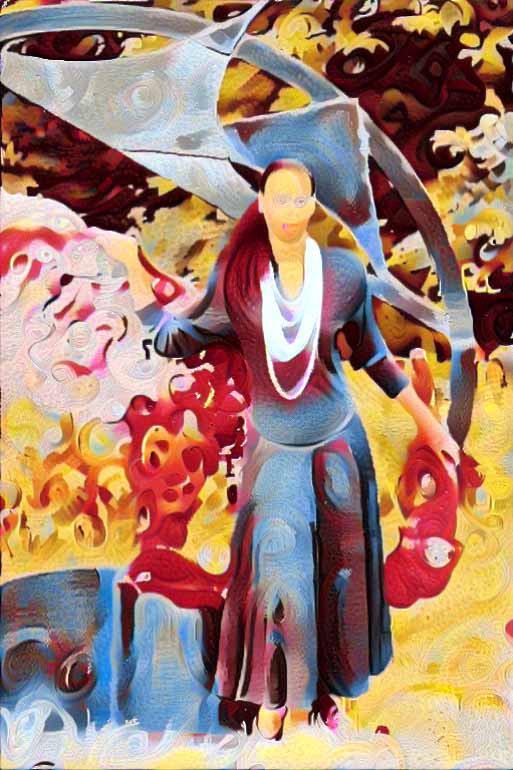}
              & \includegraphics[width=0.124\textwidth]{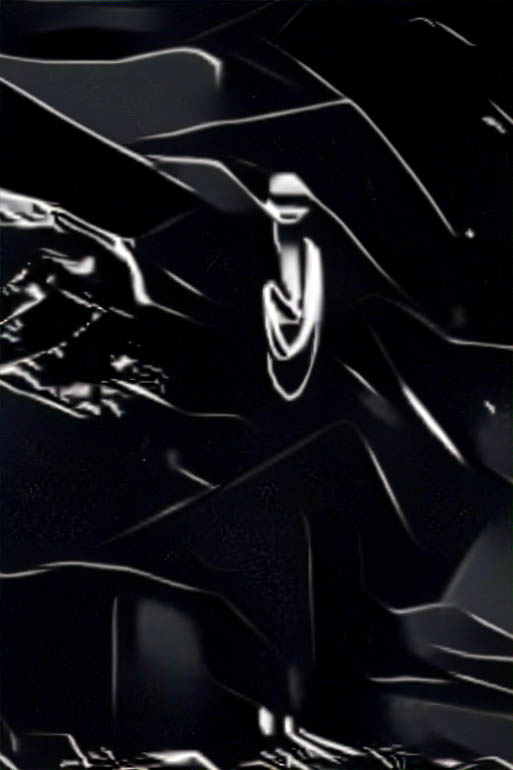}
              & \includegraphics[width=0.124\textwidth]{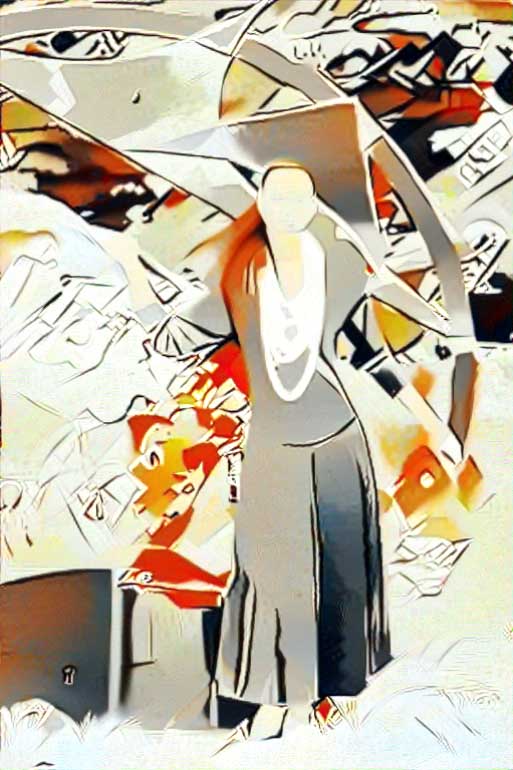}
              & \includegraphics[width=0.124\textwidth]{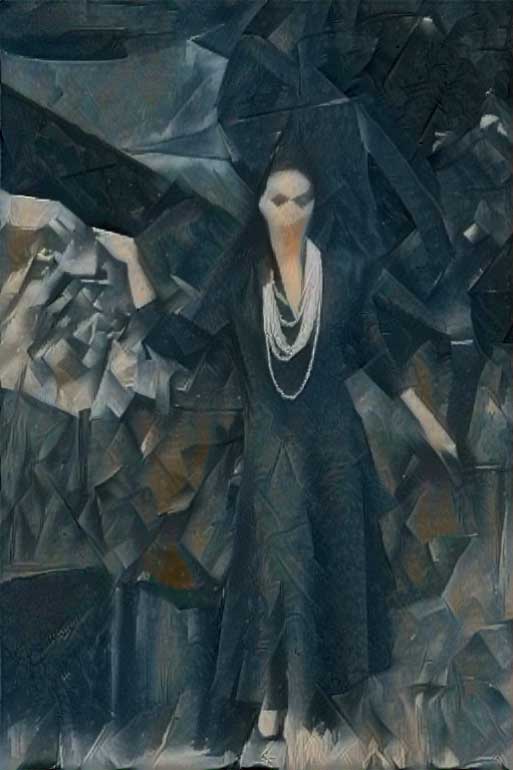}
              & \includegraphics[width=0.124\textwidth]{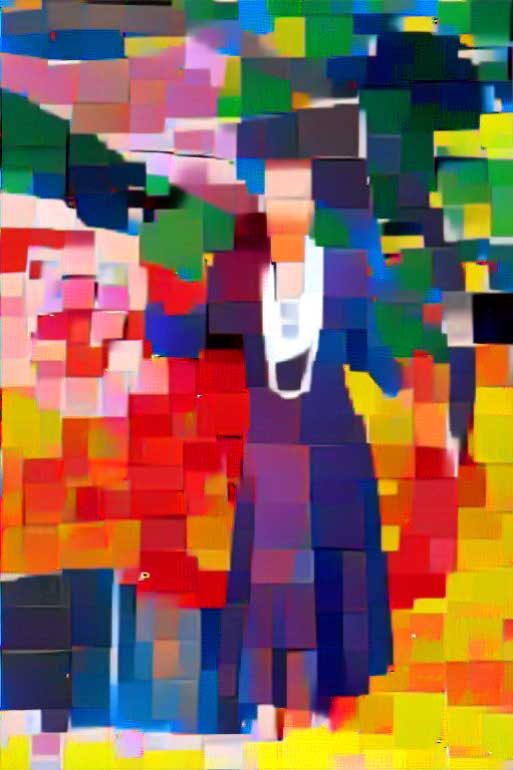}
              & \includegraphics[width=0.124\textwidth]{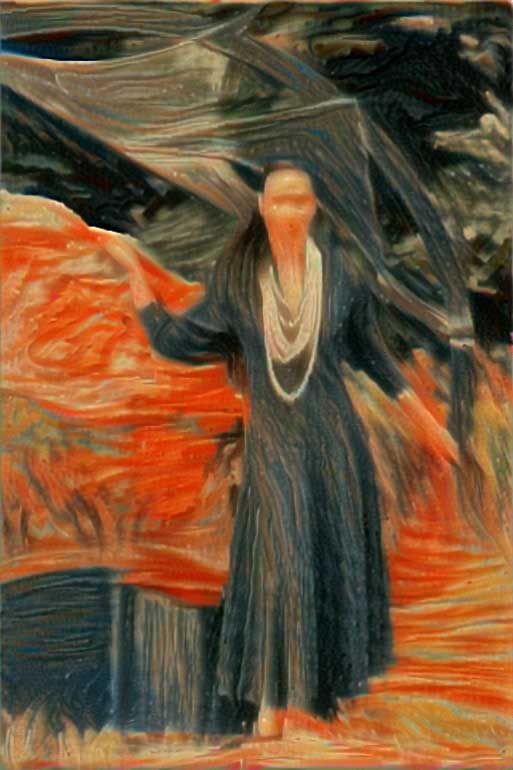}
              & \includegraphics[width=0.124\textwidth]{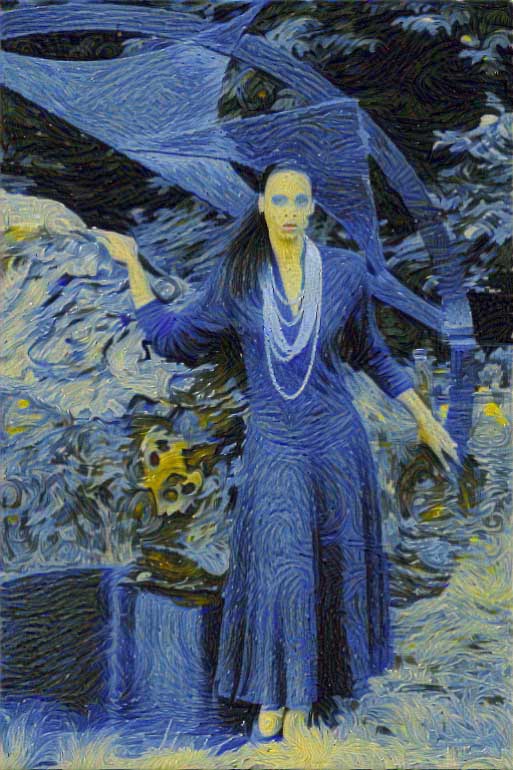}\\
              \includegraphics[width=0.124\textwidth]{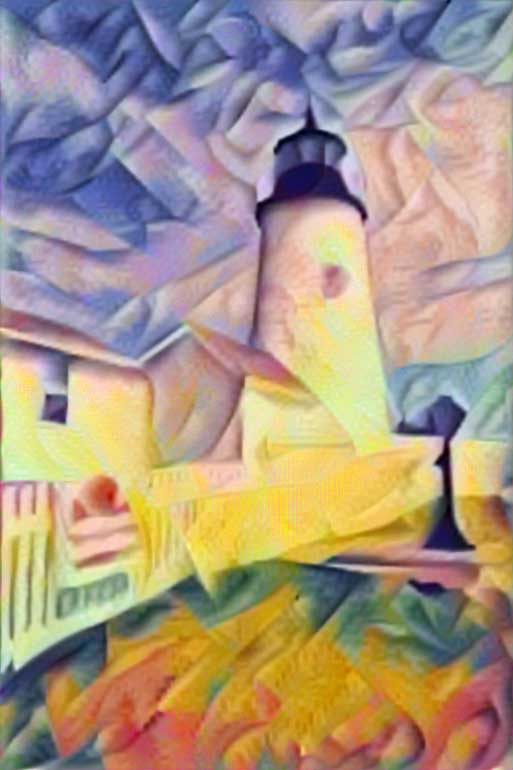}
              & \includegraphics[width=0.124\textwidth]{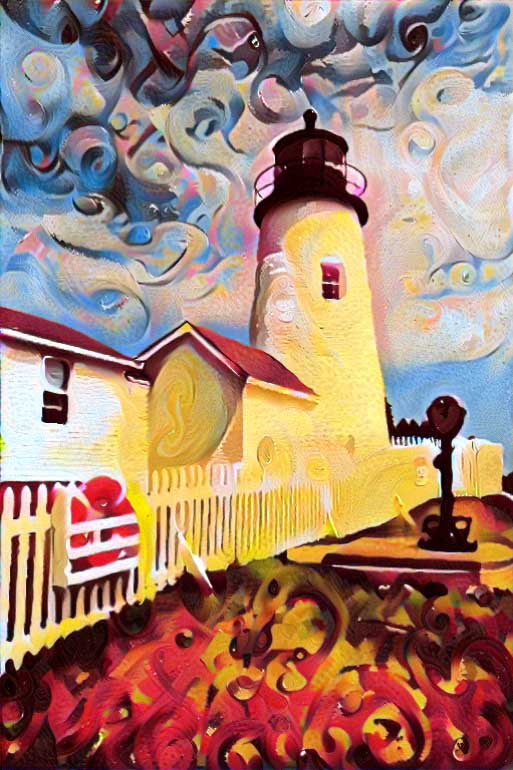}
              & \includegraphics[width=0.124\textwidth]{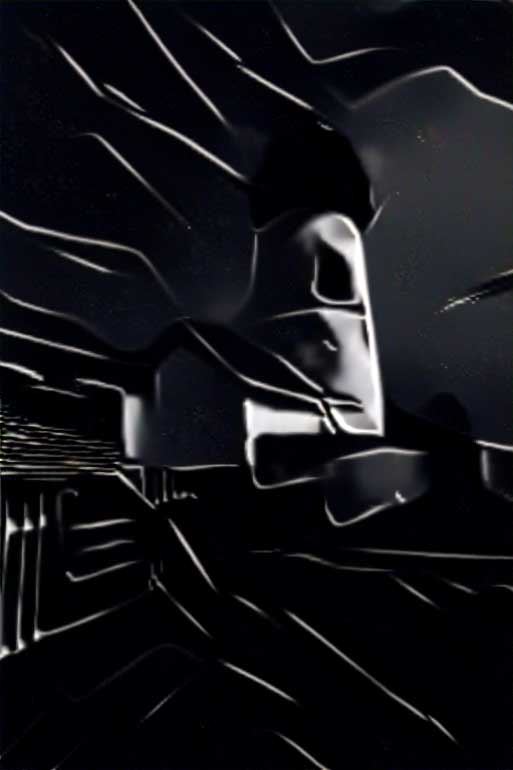}
              & \includegraphics[width=0.124\textwidth]{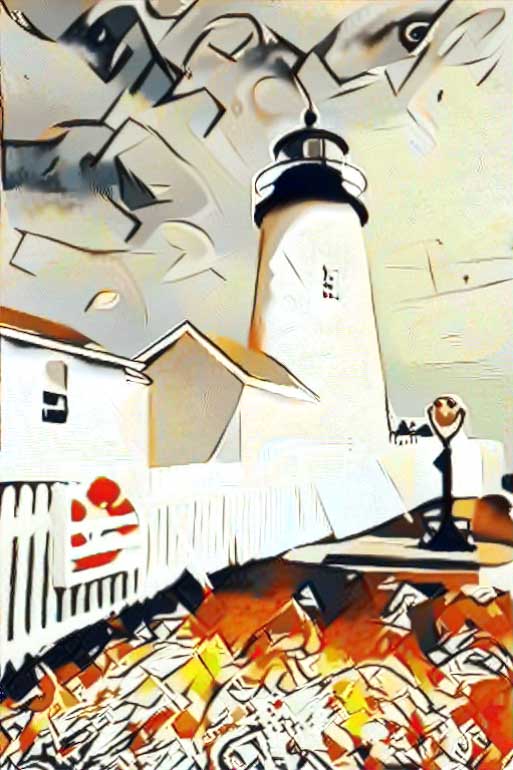}
              & \includegraphics[width=0.124\textwidth]{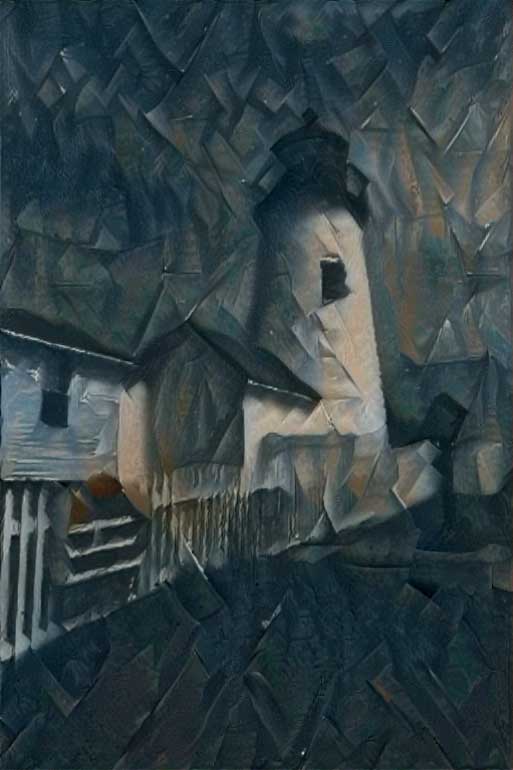}
              & \includegraphics[width=0.124\textwidth]{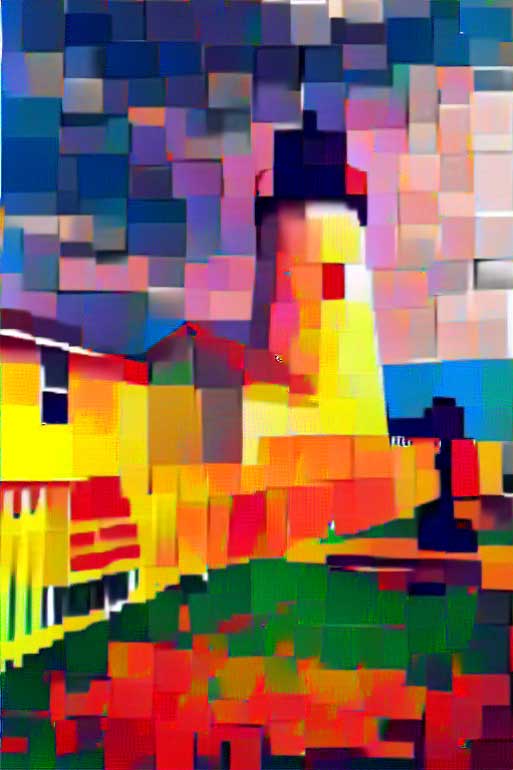}
              & \includegraphics[width=0.124\textwidth]{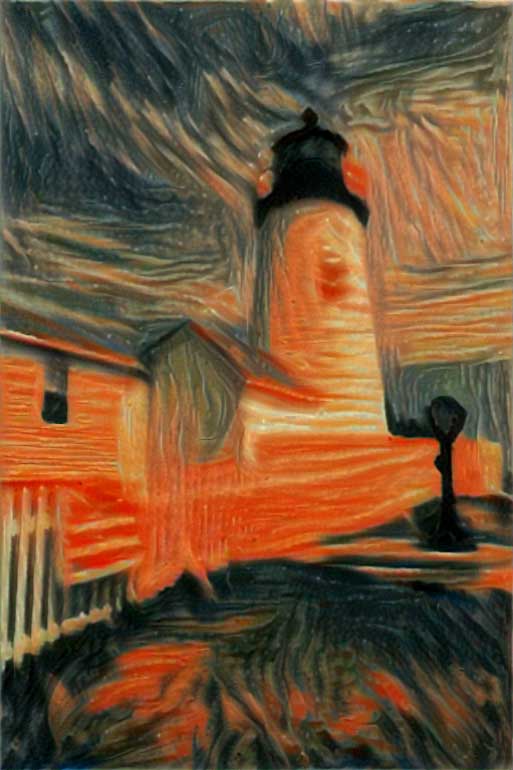}
              & \includegraphics[width=0.124\textwidth]{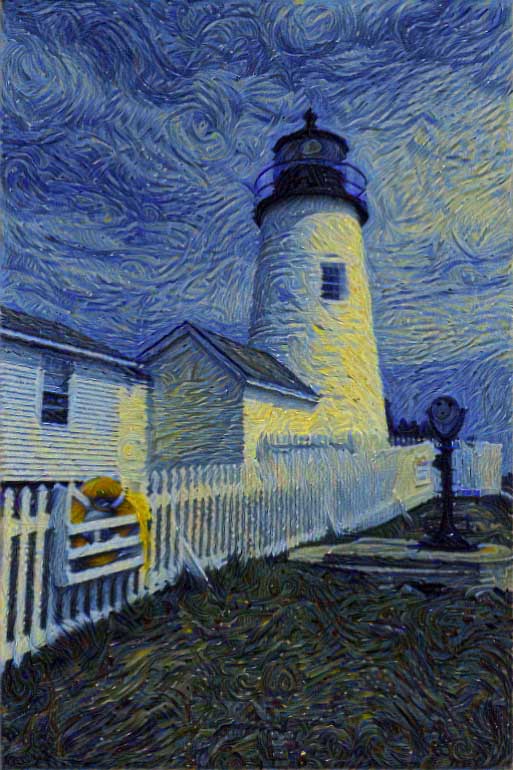}\\
        \end{tabularx}
    \end{center}
    \vspace{-1em}
    \caption{Additional qualitative style transfer results of our \ac{cmd} algorithm. All examples shown also formed part of the user study. Best viewed on screen. Please zoom in to appreciate style details.}
\end{figure}

\newpage
\section{Additional qualitative comparison}

\setlength{\tabcolsep}{1pt}

\begin{figure}[htbp]
    \begin{center}
        \begin{tabularx}{\textwidth}{*5{>{\centering\arraybackslash}X}}
              \includegraphics[width=0.195\textwidth]{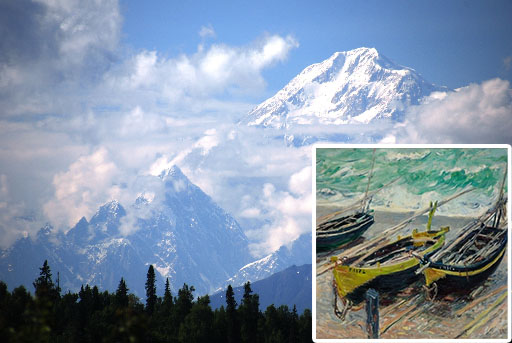}
            & \includegraphics[width=0.195\textwidth]{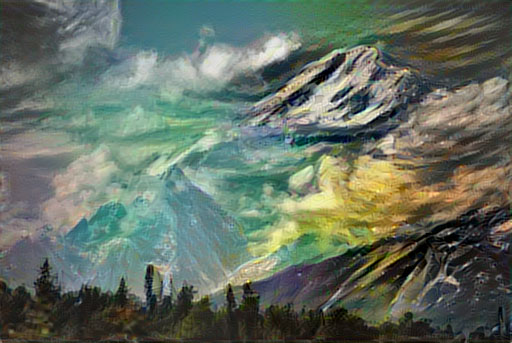} 
            & \includegraphics[width=0.195\textwidth]{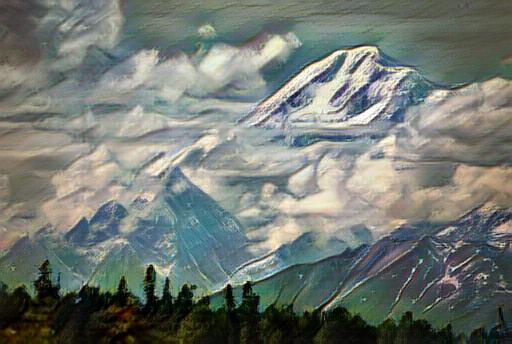} 
            & \includegraphics[width=0.195\textwidth, trim=0 4 0 4, clip]{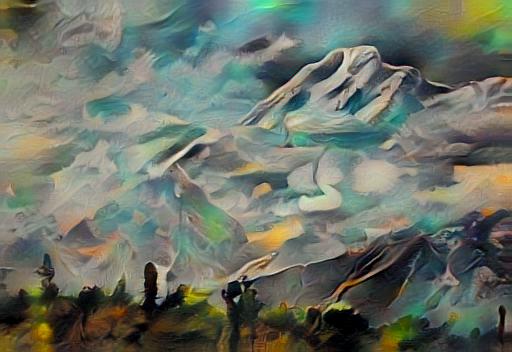} 
            & \includegraphics[width=0.195\textwidth, trim=8 0 8 0, clip]{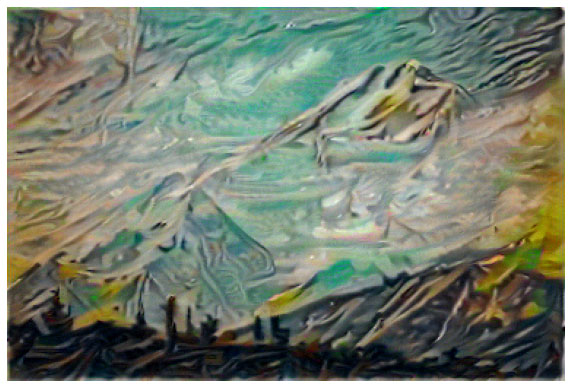}\\ 
            
              \includegraphics[width=0.195\textwidth]{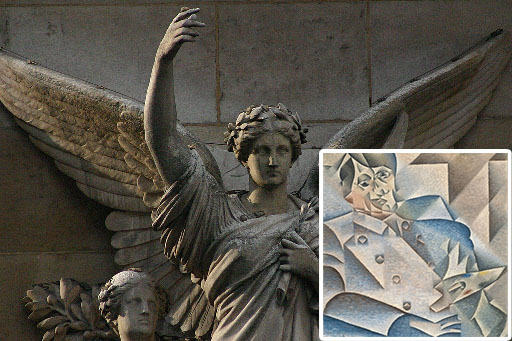}
            & \includegraphics[width=0.195\textwidth]{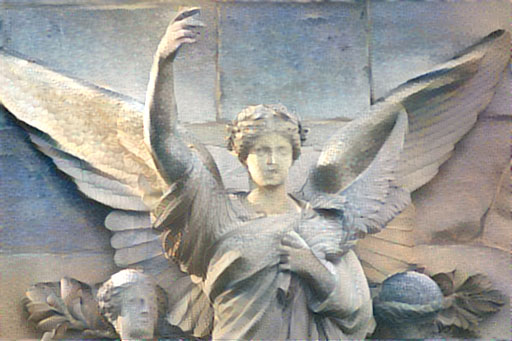}  
            & \includegraphics[width=0.195\textwidth]{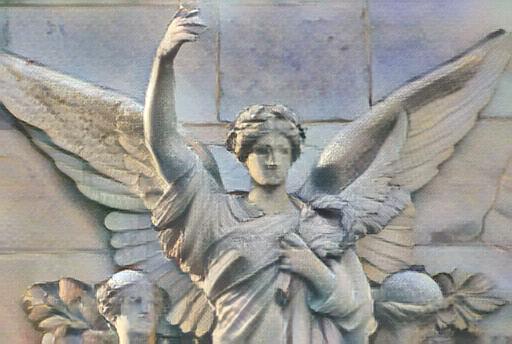}  
            & \includegraphics[width=0.195\textwidth, trim=0 4 0 4, clip]{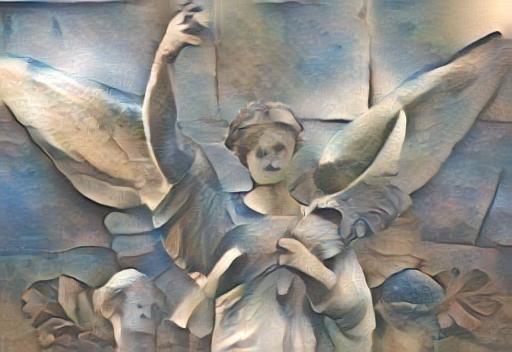}
            & \includegraphics[width=0.195\textwidth, trim=8 0 8 0, clip]{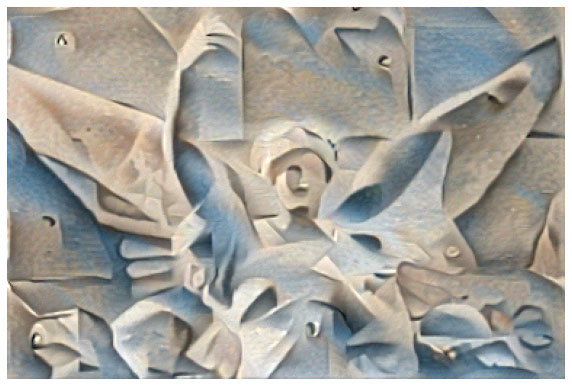}\\  
            
              \includegraphics[width=0.195\textwidth]{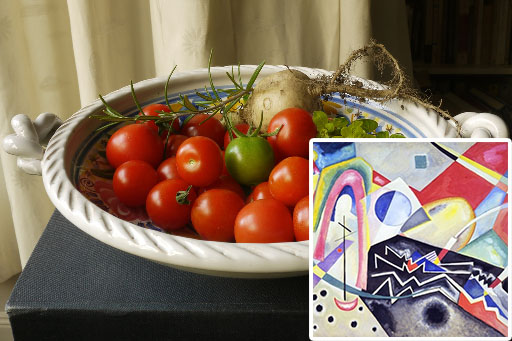} 
            & \includegraphics[width=0.195\textwidth]{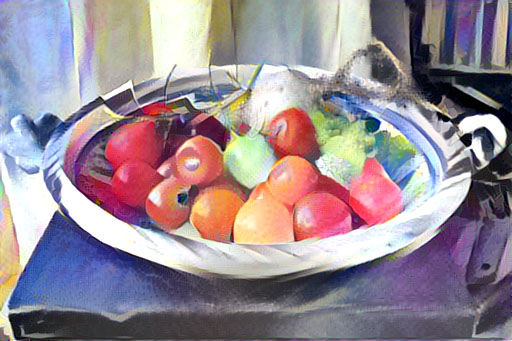} 
            & \includegraphics[width=0.195\textwidth]{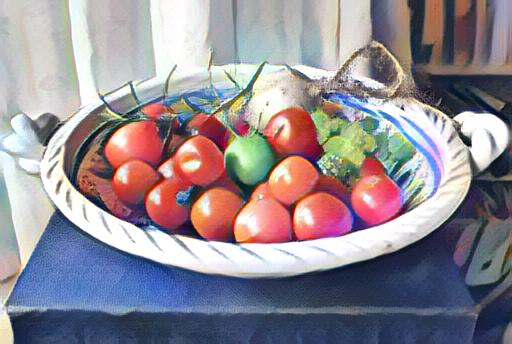} 
            & \includegraphics[width=0.195\textwidth, trim=0 4 0 4, clip]{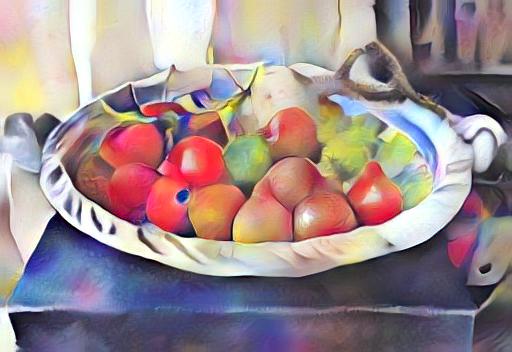} 
            & \includegraphics[width=0.195\textwidth, trim=8 0 8 0, clip]{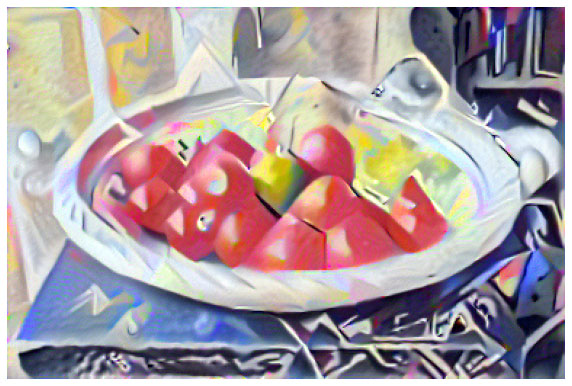}\\ 
            
              \includegraphics[width=0.195\textwidth]{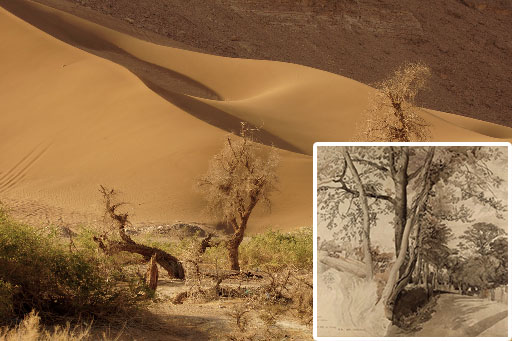}  
            & \includegraphics[width=0.195\textwidth]{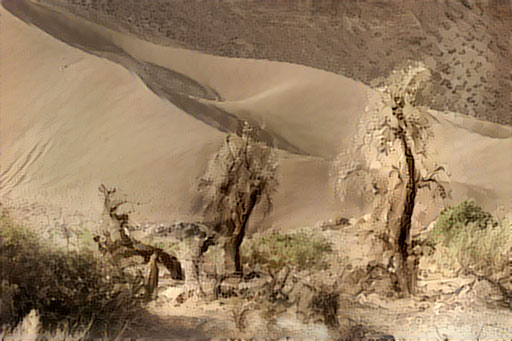}
            & \includegraphics[width=0.195\textwidth]{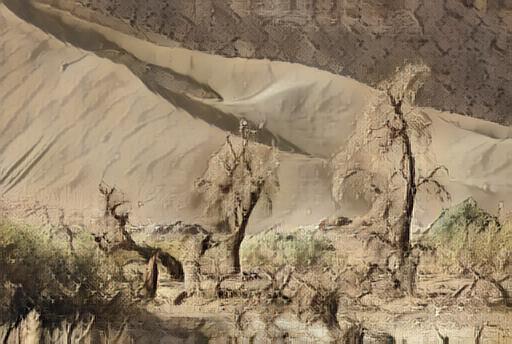}  
            & \includegraphics[width=0.195\textwidth, trim=0 4 0 4, clip]{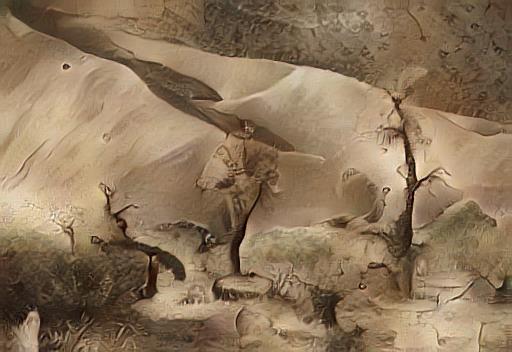}  
            & \includegraphics[width=0.195\textwidth, trim=8 0 8 0, clip]{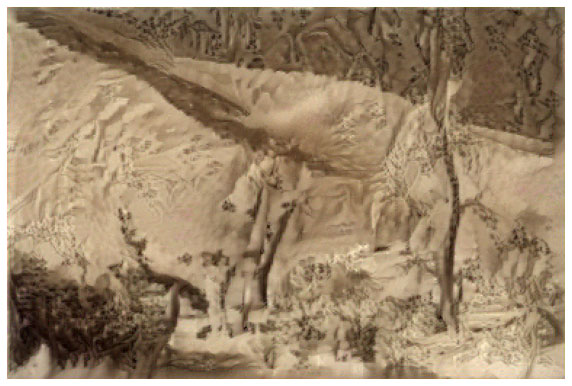}\\   
            
              \includegraphics[width=0.195\textwidth]{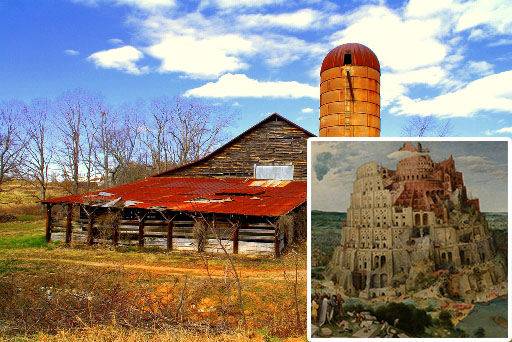} 
            & \includegraphics[width=0.195\textwidth]{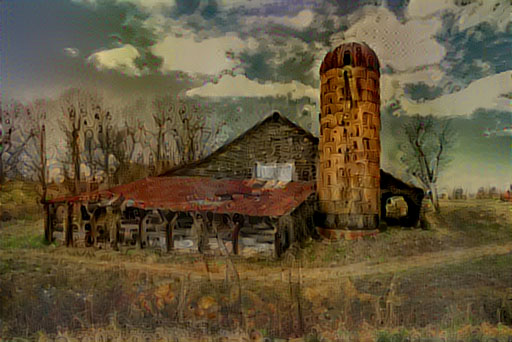} 
            & \includegraphics[width=0.195\textwidth]{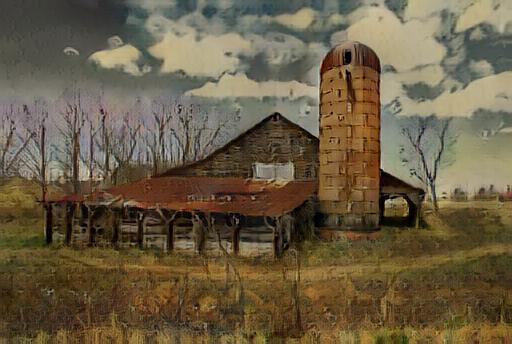} 
            & \includegraphics[width=0.195\textwidth, trim=0 4 0 4, clip]{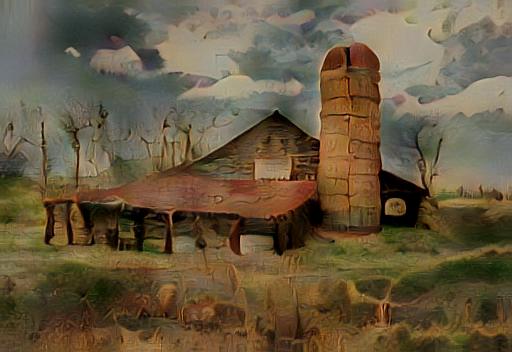} 
            & \includegraphics[width=0.195\textwidth, trim=8 0 8 0, clip]{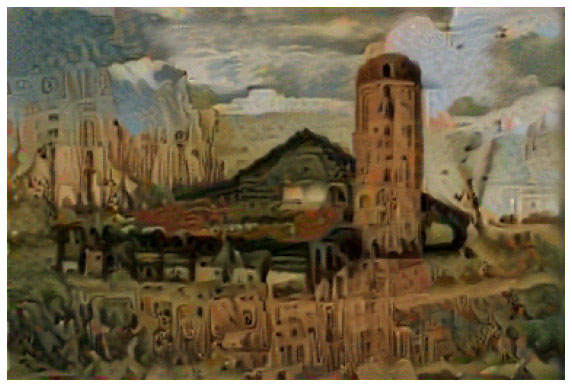}\\ 
            
              \includegraphics[width=0.195\textwidth]{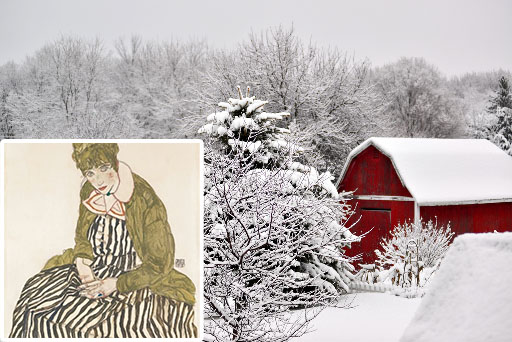}
            & \includegraphics[width=0.195\textwidth]{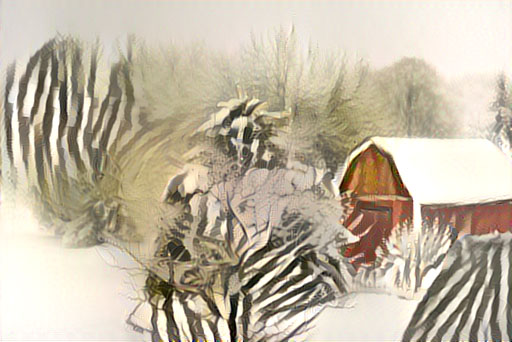} 
            & \includegraphics[width=0.195\textwidth]{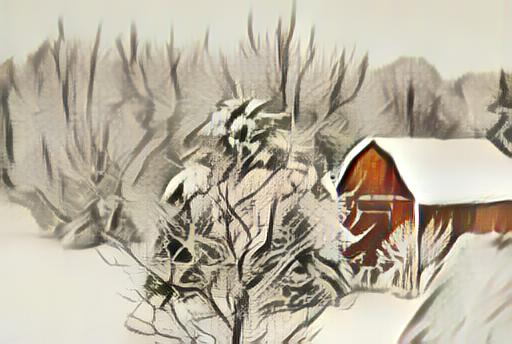}
            & \includegraphics[width=0.195\textwidth, trim=0 4 0 4, clip]{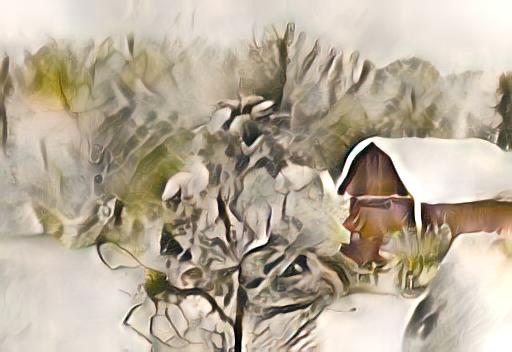} 
            & \includegraphics[width=0.195\textwidth, trim=8 0 8 0, clip]{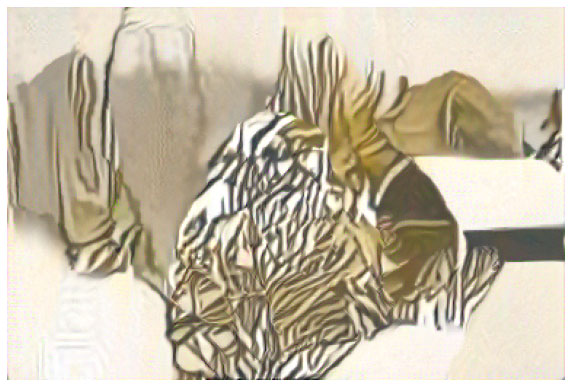}\\  
            
            \rule{0pt}{14pt}\small (a) Input 
            & \rule{0pt}{12pt}\small (b) Gatys
            & \rule{0pt}{12pt}\small (c) AdaIN  
            & \rule{0pt}{12pt}\small (d) WCT 
            & \rule{0pt}{12pt}\small (e) Ours 
        \end{tabularx}
    \end{center}
\caption{Style transfer results with our algorithm, and with one competing method per category (MM: AdaIn~\cite{huang2017arbitrary}; MMD: Gatys~\cite{gatys2016image}; OT: WCT~\cite{li2017universal}).
The displayed results for those methods were made available in \cite{jing2019neural}. Best viewed on screen. Please zoom in to appreciate style details.}
\end{figure}

\end{document}